\documentclass{article}
\PassOptionsToPackage{numbers, compress}{natbib}
\usepackage[final]{neurips_2020}

\usepackage[utf8]{inputenc} % allow utf-8 input
\usepackage[T1]{fontenc}    % use 8-bit T1 fonts
\usepackage{url}            % simple URL typesetting
\usepackage{booktabs}       % professional-quality tables
\usepackage{amsfonts}       % blackboard math symbols
\usepackage{nicefrac}       % compact symbols for 1/2, etc.
\usepackage{microtype}      % microtypography

\usepackage{scalefnt,letltxmacro}
\LetLtxMacro{\oldtextsc}{\textsc}
\renewcommand{\textsc}[1]{\oldtextsc{\scalefont{1.10}#1}}
\usepackage[acronym,smallcaps,nowarn]{glossaries}
\glsdisablehyper
\makeglossaries
\usepackage{xspace}

\usepackage{amssymb}
\usepackage{mathtools}
\usepackage{amsfonts}
\usepackage{amsmath}
\usepackage{amsthm}

\usepackage[colorlinks,linktoc=all]{hyperref}
\usepackage[all]{hypcap}
\hypersetup{citecolor=MidnightBlue}
\hypersetup{linkcolor=MidnightBlue}
\hypersetup{urlcolor=MidnightBlue}

\usepackage[nameinlink]{cleveref}
\crefname{section}{\S}{\S\S}
\Crefname{section}{\S}{\S\S}
\creflabelformat{equation}{#2\textup{#1}#3}

\usepackage[usenames,dvipsnames]{xcolor}
\definecolor{shadecolor}{gray}{0.9}

\usepackage{wrapfig}
\usepackage{subcaption}

\DeclareRobustCommand{\parhead}[1]{\textbf{#1}~}

\makeatletter
\newcommand*{\addFileDependency}[1]{% argument=file name and extension
  \typeout{(#1)}
  \@addtofilelist{#1}
  \IfFileExists{#1}{}{\typeout{No file #1.}}
}
\makeatother

\newacronym{MAP}{map}{maximum a posteriori}
\newacronym{MLE}{mle}{maximum likelihood estimation}

\newacronym{VAE}{vae}{variational autoencoder}

\newacronym{MC}{mc}{Monte Carlo}
\newacronym{MCMC}{mcmc}{Markov chain Monte Carlo}

\newacronym{VI}{vi}{variational inference}

\newacronym{ELBO}{elbo}{evidence lower bound}
\newacronym{NELBO}{nelbo}{negative evidence lower bound}

\newacronym{ELL}{ell}{expected log likelihood}
\newacronym{KL}{kl}{Kullback-Leibler}
\newacronym{MNLL}{mnll}{mean negative log likelihood}
\newacronym{MLL}{mll}{mean log likelihood}
\newacronym{ACC}{acc}{accuracy}

\newacronym{MLP}{mlp}{multilayer perceptron}
\newacronym{RELU}{ReLU}{rectified linear unit}

\newacronym{VGCN}{vgcn}{variational graph convolutional network}
\newacronym{CNN}{cnn}{convolutional neural network}
\newacronym{GCN}{gcn}{graph convolutional network}
\newacronym{GRAPHSAGE}{graphsage}{sample-and-aggregate}
\newacronym{GAT}{gat}{graph attention network}
\newacronym{LDS}{lds}{learning discrete structures}
\newacronym{EGCN}{egcn}{ensemble graph convolutional network} % AAAI paper
\newacronym{GGP}{ggp}{graph Gaussian process} %  
\newacronym{RGCN}{rgcn}{robust GCN}
\newacronym{CHEBNET}{chebnet}{Chebyshev network}
\newacronym{MCD}{mcd}{Monte Carlo dropout}
\newacronym{GP}{gp}{Gaussian process}
\newacronym{NNMF}{nnmf}{neural network matrix factorization}
\newacronym{PMF}{pmf}{probabilistic matrix factorization}
\newacronym{MONET}{monet}{mixture models network} 

\newacronym{ARD}{ard}{automatic relevance determination}

\newacronym{SVD}{svd}{singular value decomposition}

\newacronym{CF}{cf}{collaborative filtering}
\newacronym{LP}{lp}{link prediction}

\newacronym{BOW}{bow}{bag-of-words}
\newacronym{TFIDF}{tf-idf}{term frequency-inverse document frequency}
\newcommand{\adam}{\textsc{Adam}\xspace}
\newcommand{\iid}{i.i.d.\xspace}  
\newcommand{\ie}{i.e.\xspace}  
\newcommand{\eg}{e.g.\xspace}  

\newacronym{KNNG}{k-nng}{k-nearest neighbor graph} 

\newcommand{\cora}{\textsc{cora}\xspace}
\newcommand{\citeseer}{\textsc{citeseer}\xspace}
\newcommand{\polblogs}{\textsc{polblogs}\xspace}
\newcommand{\pubmed}{\textsc{pubmed}\xspace}

\newcommand{\mathbold}[1]{\ensuremath{\boldsymbol{\mathbf{#1}}}}

\newcommand{\g}{\,|\,}

\newcommand{\nestedmathbold}[1]{{\mathbold{#1}}}

\newcommand{\mbf}{\nestedmathbold{f}}

\newcommand{\mbx}{\nestedmathbold{x}}
\newcommand{\mby}{\nestedmathbold{y}}
\newcommand{\mbz}{\nestedmathbold{z}}

\newcommand{\mbA}{\nestedmathbold{A}}
\newcommand{\mbB}{\nestedmathbold{B}}

\newcommand{\mbD}{\nestedmathbold{D}}

\newcommand{\mbW}{\nestedmathbold{W}}
\newcommand{\mbX}{\nestedmathbold{X}}
\newcommand{\mbY}{\nestedmathbold{Y}}
\newcommand{\mbZ}{\nestedmathbold{Z}}

\newcommand{\mbphi}{\nestedmathbold{\phi}}
\newcommand{\mbpi}{\nestedmathbold{\pi}}

\newcommand{\mbtheta}{\nestedmathbold{\theta}}

\newcommand{\mbPi}{\nestedmathbold{\Pi}}

\newcommand{\bigO}{\mathcal{O}}

\newcommand{\IWELBO}{\textsc{iw-elbo}}
\newcommand{\LME}{\textsc{lme}}

\DeclareRobustCommand{\KL}[2]{\ensuremath{\textsc{kl}\left[#1\;\|\;#2\right]}}

\DeclareRobustCommand{\ELL}[2]{\ensuremath{\bar{\ell}_{#1} \left(#2\right)}}

\newcommand{\cL}{\mathcal{L}}

\newcommand{\cV}{\mathcal{V}}
\newcommand{\cE}{\mathcal{E}}
\newcommand{\cG}{\mathcal{G}}
 
\newcommand{\E}{\mathbb{E}}

\newcommand{\bbR}{\mathbb{R}} 

\newcommand{\Cat}{\mathrm{Cat}}
\newcommand{\Bern}{\mathrm{Bern}}
\newcommand{\BinConcrete}{\mathrm{BinConcrete}}
\newcommand{\Logistic}{\mathrm{Logistic}}
\newcommand{\Uniform}{\mathrm{Uniform}}

 \newcommand{\sigmoid}{\sigma}
 \newcommand{\elbo}{\cL_{\textsc{elbo}}}

\newcommand{\Yobs}{\mbY^{o}} % Full set of labels
\newcommand{\Ylatent}{\mbY^{u}} % Full set of labels
\newcommand{\Ahat}{\hat{\mbA}}
\newcommand{\Agiven}{\bar{\mbA}} % full observed (given) adjacency
\newcommand{\Asample}[1]{\mbA^{(#1)}}
\newcommand{\asample}[2]{A^{(#1)}_{#2}}
\newcommand{\berndiscreteprior}{\rho^{o}} % bernoulli discrete parameters prior

\newcommand{\smoothfactorzero}{\bar{\rho}_0}
\newcommand{\smoothfactorone}{\bar{\rho}_1}
\newcommand{\berndiscreteposterior}{\rho}
\newcommand{\bernrelaxedprior}{\lambda^{o}}
\newcommand{\bernrelaxedposterior}{\lambda}
\newcommand{\temprior}{\tau_{o}}
\newcommand{\temposterior}{\tau}

\begin{document}
% For affiliations
\newcommand{\footremember}[2]{%
	\thanks{#2}
	\newcounter{#1}
	\setcounter{#1}{\value{footnote}}%
}
\newcommand{\footrecall}[1]{%
	\footnotemark[\value{#1}]%
}

\title{Variational Inference for Graph Convolutional Networks in the Absence  of Graph Data and Adversarial Settings}

% The \author macro works with any number of authors. There are two commands
% used to separate the names and addresses of multiple authors: \And and \AND.
%
% Using \And between authors leaves it to LaTeX to determine where to break the
% lines. Using \AND forces a line break at that point. So, if LaTeX puts 3 of 4
% authors names on the first line, and the last on the second line, try using
% \AND instead of \And before the third author name.

\author{
Pantelis Elinas\footremember{joint}{Joint first author.}\\
\texttt{pantelis.elinas@data61.csiro.au} \\
CSIRO's Data61
\And
Edwin V.~Bonilla\footrecall{joint}\\
\texttt{edwin.bonilla@data61.csiro.au} \\
CSIRO's Data61 \& The University of Sydney
\And
Louis C.~Tiao\\
\texttt{louis.tiao@sydney.edu.au} \\
The University of Sydney \& CSIRO's Data61\\
}

%\author{%
%  Pantelis Elinas \\
%  CSIRO's Data61\\
%  \texttt{hippo@cs.cranberry-lemon.edu} \\
%  % examples of more authors
%  % \And
%  % Coauthor \\
%  % Affiliation \\
%  % Address \\
%  % \texttt{email} \\
%  % \AND
%  % Coauthor \\
%  % Affiliation \\
%  % Address \\
%  % \texttt{email} \\
%  % \And
%  % Coauthor \\
%  % Affiliation \\
%  % Address \\
%  % \texttt{email} \\
%  % \And
%  % Coauthor \\
%  % Affiliation \\
%  % Address \\
%  % \texttt{email} \\
%}
\maketitle

\begin{abstract}
We propose a framework that lifts the capabilities of graph convolutional networks (GCNs) to scenarios where no input graph is given and increases their robustness to adversarial attacks.  We formulate a joint probabilistic model that considers a prior distribution over graphs along with a GCN-based likelihood and develop a stochastic variational inference algorithm to estimate the graph posterior and the GCN parameters jointly. To address the problem of propagating gradients through latent variables drawn from discrete distributions, we use their continuous relaxations known as Concrete distributions. We show that, on real datasets, our approach can outperform state-of-the-art Bayesian and non-Bayesian graph neural network algorithms on the task of semi-supervised  classification in the absence of graph data and when the network structure is subjected to adversarial perturbations.
\end{abstract}

\section{Introduction}
\label{sec:introduction}
% CNNs powerful and successful
% Graphs are important
% spectral GCNs are neat
% We based on this to deal with the problem of noisy graphs
%
%Most previous work on GCNs was on generalizing convnets to graph structured data. one of the motivations is that for data of dimension D., deep learning strategies usually employ fully-connected layers, which have $O(D^2)$ parameters. Hence, clever regularization schemes are required such as weight decay, dropout, etc. Using a generalization of convents would help here, was it would reduce significantly the number of parameters to estimate while maintaning the capacity to extract useful statistics from the data. The problem is defining CNN operations on the grid ... bla bla bla.
%
%
% graph constructed over gratures
%\cite{bruna2014spectral}: pioneering work on graph nn: propose spatial strategy and spectral strategy to graph CNNs. Propose the spline smoothing
%\cite{henaff2015deep}: extend the above to large-scale classification and estimate the graph using the data
%\cite{defferrard2016convolutional} more theoretical foundation of graph CNNs and efficient implenetation through Chebishev polynomials
% network analysis and discovery
% social networks 
% cite Adams 

Graphs represent the elements of a system and their relationships as a set of nodes and edges, respectively. By exploiting the inter-dependencies of these elements, many applications of machine learning have achieved significant success, for example in the areas of social networks \cite{hamilton2017inductive}, document classification \cite{kipf2016semi,velickovic2018graph} and bioinformatics \cite{fout2017protein}. 
%Recent notable theoretical and practical insights in graph-based analysis  \cite{hammond2011wavelets,bruna2014spectral,henaff2015deep,kipf2016semi} along with breakthroughs in deep learning and big data processing, have reignited research interest in problems such as network analysis and discovery \cite{kipf2018neural}, relational learning \cite{schlichtkrull2018modeling}, semi-supervised learning \cite{kipf2016semi,velickovic2018graph}, generative models for graphs \cite{de2018molgan,lloyd2012random}, and, more generally, graph-based inductive inference  \cite{hamilton2017inductive}. 
In particular, motivated by the incredible success of \acrlongpl{CNN} \citep[\acrshortpl{CNN},][]{lecun1998gradient} on regular-grid data, researchers have generalized some of their fundamental properties (such as their ability to learn local stationary structures efficiently) to graph-structured data \cite{bruna2014spectral,henaff2015deep,defferrard2016convolutional}. These approaches mainly focused on exploiting feature dependencies explicitly defined by a graph in an analogous way to how \glspl{CNN} model  long-range correlations through local interactions across pixels in an image. The seminal work by \cite{kipf2016semi} leveraged these ideas to model \emph{dependencies across instances} (instead of features) to be able to incorporate  knowledge of the instances' relationships in a semi-supervised learning setting, going beyond the standard \iid assumption.

In this work we focus precisely on the problem of semi-supervised classification based on the method developed in \cite{kipf2016semi}, which is now commonly referred to as \acrfullpl{GCN}. These networks can be seen as a first-order approximation of the spectral graph convolutional networks developed by \cite{defferrard2016convolutional}, which itself built upon the pioneering work of \cite{bruna2014spectral,henaff2015deep}. The great popularity of \glspl{GCN} is mainly due to their practical performance as, at the time it was published, it outperformed related methods by a significant margin. Another practical advantage of using \glspl{GCN} is their relatively simple propagation rule, which does not require expensive operations such as eigen-decomposition. 

However, one of the main  assumptions underlying \glspl{GCN}  is that the given graph is helpful for the task at hand and that the corresponding edges are highly reliable. This is generally not true in practical applications, as the given graph may be (i)  noisy, (ii) loosely related to the classification problem of interest or (iii) built in an ad hoc basis using, \eg, side information. 
Although these settings have been addressed  by previous work independently \citep[see e.g.][]{lds-2019,robustgcn2019}, in practice, it is difficult to incorporate this type of uncertainty over the graph using the original \gls{GCN} framework in a principled way and the performance of the method degrades significantly with increasingly noisy graphs. 

Consequently, in this paper we propose a framework that lifts \gls{GCN}'s capabilities 
% Therefore, in this paper we extend \gls{GCN} capabilities 
to handle the more challenging cases of learning in the absence of an input graph and dealing with  highly-effective adversarial perturbations such as those proposed recently by \cite{pmlr-v97-bojchevski19a}. We achieve this by formulating a joint probabilistic model that places a prior over the graph structure, as given by the adjacency matrix, along with a GCN-based conditional likelihood. This, however, poses significant inference challenges as estimating the posterior over the adjacency matrix  under a highly non-linear likelihood (as given by  the \gls{GCN}'s output) is intractable. 
 
Thus, to estimate the graph posterior we resort to approximations via  stochastic \gls{VI}. Nevertheless, even in the approximate inference world, carrying out posterior estimation over a very large discrete combinatorial space can prove extremely hard. We adopt a simple but effective relaxation over both the prior and the approximate posterior using Concrete distributions \cite{maddison2016concrete}, which allows us to propagate gradients in the corresponding stochastic computational graph.  
%To the best of our knowledge, we are the first to develop an efficient variational Bayesian approach that incorporates uncertainty information about the given graph directly and to use in the absence of graph data and in adversarial settings.
Our experiments show that we can outperform state-of-the-art Bayesian and non-Bayesian graph neural network algorithms in the task of semi-supervised classification (i) in the absence of graph data;  (ii) when the network structure is subjected to adversarial perturbations and (iii) when considering the ground truth graphs. Our results and analyses indicate that our framework does indeed learn new graph representations by turning on/off connections so as to improve performance on the given task. 

% We can put this back in, in case the lowrank works!
% Nevertheless, a crucial remaining challenge is how to parameterize the posterior over $\bigO(N^2)$ parameters, where $N$ is the number of nodes in the graph.  We propose a simple but effective parameterization inspired by low-rank matrix factorization, where we assume  underlying continuous latent variables for each row/column of the posterior parameters over the adjacency matrix. 

\subsection{Related work}
\label{sub:related_work}
Most graph neural network frameworks can be seen from a more general perspective under the unifying \acrlong{MONET} \citep[\acrshort{MONET},][]{monti2017geometric}. 
%Before \glspl{GCN}, the problem of semi-supervised classification had been traditionally tackled via optimization of  a supervised loss along with  graph Laplacian regularization \cite{zhu2003semi,belkin2006manifold,weston2012deep} or via graph embeddings  \cite{perozziDeepwalk2014,tangline2015,yang2016revisiting}. \glspl{GCN}, as proposed by \cite{kipf2016semi}, seem to have been the first to use spectral convolutional networks for this problem. However, these approaches are inherently transductive and do not generalize to unseen nodes. Hence, improvements to \glspl{GCN} include inductive generalizations \cite{hamilton2017inductive} and `attention' mechanisms \cite{velickovic2018graph}. In fact, \glspl{GAT} as proposed by  \cite{hamilton2017inductive} can be somewhat more resilient to noisy settings. For more details of graph neural network approaches the reader is referred to the excellent related work in \cite{velickovic2018graph} and, more generally, to the surveys in \cite{zhou2018graph,wu2019comprehensive}.
For details of graph neural network approaches the reader is referred to the excellent related work in \cite{velickovic2018graph} and, more generally, to the surveys in \cite{zhou2018graph,wu2019comprehensive}.
%
%\textbf{Uncertain graphs.} Previous works have looked at learning settings with uncertain graphs.
%For example, 
With regards to uncertain graphs, \cite{boldi2012injecting} propose a  graph-anonymization technique that injects uncertainty in the existence of edges of the graph. \cite{dallachiesa2014node} propose a method that models the probability of a node  belonging to a particular class as a function of the uncertainty in the edges related to that node. However, such an approach does not build upon state-of-the-art graph networks nor develop a fully coherent probabilistic model over the parameters of the network. Following a different methodology, \cite{hu2017embedding}  deal with the problem of uncertain graphs via embeddings by constructing a proximity matrix given the uncertain graph and applying matrix factorization to get the embedded representation. The embeddings are then used in supervised learning tasks. Unlike our method, this is a two-step procedure  where the prediction task is separate from  uncertainty modeling and representation learning. 

%First and foremost, despite their method being referred to as Bayesian, their model is an odd choice as their \emph{true posterior} parameterization over the graph does not depend on the observed data (neither features or labels). Second, their algorithm ends up training $N_G$ \glspl{GCN} using \gls{MCD}, were $N_G$ is the number of graph samples. This is significantly more computationally costly than our method. Third, incorporating prior information over the graph parameters in their approach is less natural than in our method, as their random graph model is based on mixed membership stochastic block models.  Last, unlike ours, their main focus is in the low-labeled-data regime, although they also present some results in an adversarial setting. 
% \textbf{Probabilistic approaches.} 
In terms of probabilistic approaches, 
% Having a similar motivation to ours, 
\cite{zhang2018bayesian} propose a %probabilistic 
model for \glspl{GCN} where the graph is considered as a  realization of  mixed membership stochastic block models \cite{airoldi2008mixed}. However, despite their method being referred to as Bayesian, it parameterizes the \emph{true posterior}  over the graph directly and this posterior is not dependent  on the observed data (neither features or labels). Thus, it can be seen more like an ensemble \gls{GCN}. 
%Furthermore, unlike ours, their main focus is on the low-labeled-data regime, although they also present some results in an adversarial setting. 
Targeting the low-labeled-data regime, \cite{ng2018bayesian} propose a method for semi-supervised classification with \gls{GP} priors, where the parameters of a robust-max likelihood for a  node are given by the  average of the \gls{GP} values over its 1-hop neighborhood. Their results indicate that their method can outperform \glspl{GCN} in active learning settings. 

Concerning graph structure learning, similarly to our work, \cite{kipf2018neural}  use variational autoencoders during inference  to learn this structure. However, their focus is very different to ours as they  address the problem of learning the interactions between components in a dynamical system given their trajectories (i.e., the entities evolve over time) in an unsupervised way. % Such problems are also of interest in, e.g., causality and computational neuroscience. 
More recently, \cite{lds-2019} develop a method for learning the graph structure using a generative model and estimate its parameters via bilevel optimization. However, unlike our approach, their method is not Bayesian and it does not allow for either the incorporation of prior knowledge or estimation of the full posterior over the graph. To elaborate on this point, its initial edge probabilities use a deterministic distribution (see their Algorithm 1) and do not play an explicit role in the objective function being optimized, i.e.~there is no  prior constraining the search-space over graphs. Using different initial probabilities will not achieve a similar effect to that obtained in our joint  probabilistic framework.
Finally, \cite{kipf2016variational} propose the variational graph autoencoder, a probabilistic framework  that % also 
learns latent representations for graphs but, unlike our work, is designed for the task of link prediction. 

% \ebnote{Many `propose' here. Can change this?}

\section{Bayesian graph convolution models}
\label{sec:spectral_graph_convolution_models}
%
%\todo[inline]{We need both a schematic description and a graphical model representation of our approach}
Let $\mbX \in \bbR^{N \times D}$ be a set of $D$-dimensional features representing $N$ instances and $\mbY = \{ \mby_n\}$ be their corresponding labels,  some of which  are observed and others unobserved and $\mby_n \in \{0,1\}^C$ is one-hot-encoded. The goal of semi-supervised classification is to leverage the labeled and unlabeled data in order to predict the unobserved labels. In this paper we are interested in doing so by explicitly exploiting the dependencies among datapoints as given by an undirected graph $\cG = (\cV, \cE)$  with $N$ nodes $v_i \in \cV$, edges $(v_i, v_j) \in \cE$ and binary adjacency matrix $\mbA \in \{0,1\}^{N \times N}$. 

\parhead{GCN's basic propagation rule.} To do this, we consider the popular \gls{GCN} models \cite{kipf2016semi}, which can be seen as first-order approximations to more general (but computationally costly) spectral graph convolutional networks \cite{defferrard2016convolutional}.  For a signal  $\mbX$, \cite{kipf2016semi} showed that  one can write a convolved signal matrix as $\tilde{\mbD}^{-\frac{1}{2}} \tilde{\mbA} \tilde{\mbD}^{-\frac{1}{2}} \mbX \mbW$, where $\mbW$ is a matrix of filter parameters,  $\tilde{\mbA} = \mbA + \mathbf{I}_N$ is the adjacency matrix of the graph augmented with self-loops and $\tilde{\mbD}$ is  the corresponding  degree matrix with $\tilde{\mathbf{D}}_{ii} = \sum_{j=1}^N \tilde{\mbA}_{ij}$. This convolved signal matrix constitutes the basic operation in \glspl{GCN}. 

\parhead{Composition of graph convolutions.} 
Thus, we can define compositions  of these approximate spectral graph convolutions %$\mbf^{(l)}(\mathbf{X}, \mathbf{A})$ 
 by the recurrence relation,
$
%\begin{equation}
%\begin{split}
\mbf^{(0)}(\mathbf{X}, \mathbf{A}) 
 = \mathbf{X}
 $
, 
$
\mbf^{(l+1)}(\mathbf{X}, \mathbf{A}) 
 = h^{(l+1)} \left ( 
\Ahat
\mbf^{(l)}(\mathbf{X}, \mathbf{A})
\mathbf{W}_{l} 
\right )    ,
%\end{split}
%\end{equation}
$
where $\Ahat \equiv \tilde{\mathbf{D}}^{-\frac{1}{2}} \tilde{\mbA} \tilde{\mathbf{D}}^{-\frac{1}{2}}$,  $\tilde{\mbA}$ and $\tilde{\mbD}$ defined as above and $h^{(l)}(\cdot)$ is a nonlinear activation function of the $l$-th 
layer, typically the element-wise \gls{RELU}, $\mathtt{relu}(\cdot) =
\max(0, \cdot)$.
\gls{GCN} uses these types of compositions to define a neural network architecture for semi-supervised classification, where the activation for the final layer $h^{(L)}(\cdot)$ is the row-wise softmax function. Each layer is parameterized by $\mathbf{W}_l$, a $Q^{(l)} \times Q^{(l+1)}$ matrix of weights, where $Q^{(l)}$ is the number of hidden units for layer $l$, with  $Q^{(0)} = D$ and $Q^{(L)} = C$.
We will denote the \gls{GCN} parameters with $\mbtheta = \{ \mathbf{W}_{l} \}_{l=1}^L$.% and are trained, in the original method, by minimization of the cross-entropy error. 

\parhead{Two-layer \gls{GCN}.} In this paper we will focus on two-layer architectures, hence the output of the \gls{GCN} at the last layer, \ie  $\mbPi \equiv \mbf^{(L)}(\mathbf{X}, \mathbf{A})$ is given by:
\begin{equation}
\label{eq:gcn-two-layers}
\mbPi = \mathtt{softmax}
\left ( 
%\tilde{\mathbf{D}}^{-\frac{1}{2}}
%\tilde{\mbA} 
%\tilde{\mathbf{D}}^{-\frac{1}{2}} 
\Ahat 
\mathtt{relu}
\left ( 
%\tilde{\mathbf{D}}^{-\frac{1}{2}}
%\tilde{\mbA} 
%\tilde{\mathbf{D}}^{-\frac{1}{2}} 
\Ahat
\mathbf{X}
\mathbf{W}_{0} 
\right )
\mathbf{W}_{1} 
\right ),
\end{equation}
where $\mbPi$ is an $N \times C$ matrix of probabilities for all nodes and classes.

When the given graph is highly reliable, \glspl{GCN} trained with the proposed cross-entropy minimization method can yield state-of-the-art classification results  \cite{kipf2016semi}. However, one would like to lift \gls{GCN} capabilities to scenarios where there is \emph{no input graph} or make them robust to cases when the graph has been perturbed \emph{adversarially}. 
% \gls{GCN}'s performance  degrades significantly.  %and we require encoding our beliefs about the given graph in a principled way as well as having alternative methods to estimate its parameters. 
In order to address these problems, we propose a joint probabilistic model  that considers the graph parameters as random variables and develop a stochastic variational  inference algorithm to estimate the posterior over these parameters. This posterior is then used in conjunction with the \gls{GCN} parameters for making predictions over the unlabeled instances. 
\subsection{Likelihood}
\label{sub:observation_model} 
Our likelihood model assumes that, conditioned on all the features and the graph adjacency matrix, the observed labels $\Yobs$ are conditionally independent, \ie,
\begin{equation}
\label{eq:likelihood}
%\begin{split}
p_{\mbtheta}(\Yobs \g \mbX, \mbA) 
%&
= \prod_{\mby_n \in \Yobs} p_{\mbtheta}(\mby_n \g \mbX, \mbA) \quad \text{ with } \quad
p_{\mbtheta}(\mby_n \g \mbX, \mbA)
%&
= \Cat(\mby_n \g \mbpi_n),
%\end{split}
\end{equation}
%
%\begin{equation}
%p_{\mbtheta}(\mby_n \g \mbX, \mbA)
%= \mathrm{Cat}(\mby_n \g \mbpi) 
%\end{equation}
%\begin{equation}
%\log p_{\mbtheta}(\mby_n \g \mbX, \mbA) = \sum_{k=1}^K y_{nk} \log \pi_{nk}
%\end{equation}
where $\Cat(\mby_n | \mbpi_n)$ denotes a Categorical distribution over $\mby_n$ with parameters $\mbpi_n$  being the $n$-th row of the $N \times C$ probability matrix $\mbPi$ obtained from a \gls{GCN} with $L$ layers, \ie, $\mbPi = \mbf^{(L)}(\mathbf{X}, \mathbf{A})$. As before, $\mbtheta$ denotes the \gls{GCN}'s weight parameters. One of the fundamental differences of our approach with standard \glspl{GCN} is that we consider a prior over graphs that is constructed using the observed (but potentially noisy or unreliable) adjacency matrix.
\subsection{Prior over graphs}
\label{sub:prior}
We consider % Erd\H{o}s-R\'{e}nyi 
% EB: If p is different according to whether A_{ij}=1, then we don't really have a ER random graph prior
random graph priors of the form
\begin{equation}
\label{eq:prior}
  p(\mbA) = \prod_{ij} p(A_{ij}),  \text{ with }  p(A_{ij}) = \Bern(A_{ij} \g \berndiscreteprior_{ij}) ,
\end{equation}
where $\Bern(A_{ij} \g \berndiscreteprior_{ij})$ is a Bernoulli distribution over $A_{ij}$ with parameter $\berndiscreteprior_{ij}$. Our prior is constructed given an observed (auxiliary) graph $\Agiven$ but, for simplicity in the notation, we  omit this conditioning here and in all related distributions.  This prior can be constructed in various ways so as to encode our beliefs about the graph structure and about how much this structure should be trusted a priori for our semi-supervised classification task. For example, we have found that the construction $\berndiscreteprior_{ij} = \smoothfactorone \Agiven_{ij}  + \smoothfactorzero ( 1 - \Agiven_{ij} )$, with  $0 < \smoothfactorone,\smoothfactorzero< 1$ being hyper-parameters, works very well in practice as it gives just enough flexibility to encode our degree of belief on the absence and presence of links separately. 
% Alternatively, we can also consider a hierarchical approach where we place a prior over the parameters of the Bernoulli distribution, for example a Beta distribution. However, we have found experimentally that there are not additional benefits from introducing this hierarchy in the prior. 

In the adversarial setting, \ie, when the adjacency matrix of the given graph is altered through adversarial perturbations, $\Agiven$ is simply given by the perturbed matrix. In the case when no input graph is provided, $\Agiven$ can be obtained through a \gls{KNNG}. In other words, given a distance function $d(\cdot, \cdot)$, $\Agiven_{i,j}=1$ iff $d(\mbx_i, \mbx_j)$ is among the $k$ smallest distances from $\mbx_i$ to all other instances % ${\mbx_k}$
and $\Agiven_{i,j}=0$ otherwise.

\section{Graph structure inference}
\label{sec:latent_network_structure_inference}
Our goal is to carry out joint inference over the \gls{GCN} parameters and the graph structure as given by the adjacency matrix. Since the main additional component of our approach is to consider a prior over the adjacency matrix, we focus on the estimation of the posterior over this matrix given the observed data, i.e.~$p(\mbA \g \mbX, \Yobs) \propto p_{\mbtheta}(\Yobs \g \mbX, \mbA) p(\mbA)$ where the likelihood and prior terms are given by  \cref{eq:likelihood,eq:prior}, respectively. 

Computation of this posterior is analytically intractable due to the highly non-linear nature of the likelihood so we resort to approximate posterior inference methods. Given the high-dimensional nature of the posterior over $\mbA$, we focus on compact representations of the posterior via \gls{VI} \cite{jordan1999introduction}. Generally, \gls{VI} methods entail the definition of an approximate posterior (variational) distribution and the estimation of its parameters via the maximization of the so-called \acrfull{ELBO}, a procedure that is known to be equivalent to minimizing the \gls{KL}  divergence between the approximate and true posterior distributions. 
\subsection{Variational distribution: free parameterization and continuous relaxations}
Similarly to the prior definition, our approximate posterior is of the form
\begin{equation}
%\begin{split}
\label{eq:main-var-distro}
q_{\mbphi}(\mbA) = \prod_{ij} q_\mbphi(A_{ij}),  \text{ with } \quad
 q_{\mbphi}(A_{ij}) = \Bern(A_{ij} \g \berndiscreteposterior_{ij}),  \berndiscreteposterior_{ij}>0,
%\end{split}
\end{equation}
where, henceforth, we use $\mbphi$ to denote all the parameters of the variational posterior. In the case where $\berndiscreteposterior_{ij}$ are free parameters then $\mbphi = \{\berndiscreteposterior_{ij}\}$. We refer to this approach as the \emph{free} parameterization.  The variational distribution defined in \cref{eq:main-var-distro} naturally models the discrete nature of the adjacency matrix $\mbA$. Our goal is then to estimate the parameters $\mbphi$ of the posterior $q_{\mbphi}(\mbA)$ via maximization of the \gls{ELBO}. For this purpose we can use the so-called {score function} method \cite{ranganath2014black}, which provides an unbiased estimator of  the gradient of an expectation of a function using \gls{MC} samples. However, it is now widely accepted that, because of its generality, the score function estimator can suffer from high variance \cite{ranganath2016hierarchical}. 
 
Therefore, as an alternative to the score function estimator, we can use the so-called re-parameterization trick \cite{kingma2013auto,rezende2014stochastic}, which generally exhibits lower variance. Unfortunately, the re-parameterization trick is not applicable to discrete distributions so we need to resort to continuous relaxations. In this work we use Concrete distributions as proposed by  \cite{jang2016categorical,maddison2016concrete}. In particular, we  denote our binary Concrete posterior distribution with location parameters $\bernrelaxedposterior_{ij} > 0 $ and temperature $\temposterior > 0$ as $q_{\mbphi}(A_{ij}) = \BinConcrete(A_{ij} \g \bernrelaxedposterior_{ij}, \temposterior)$. Analogously, as discussed in \cite{maddison2016concrete}, in order to maintain a lower bound during variational inference we also relax our prior so that $p(A_{ij}) = \BinConcrete(A_{ij} \g \bernrelaxedprior_{ij}, \temprior)$. 
 In this case the variational parameters are the parameters of the Concrete distribution  $\mbphi = \{\bernrelaxedposterior_{ij}\}$.
 
 Our experiments in \cref{sec:experiments} focus on freely-parameterized variational posteriors using continuous relaxations via Concrete distributions. However, we have also analyzed the performance of discrete approaches along with low-rank parameterizations such as those used by \cite{kipf2016variational}. These analyses, detailed in the supplement,   show that our approach is superior and can be explained by recent results regarding the severe limitations of low-rank representations of graphs \cite{Seshadhri5631}. 

\subsection{Evidence lower bound}
It is easy to show that we can write the \gls{ELBO} as
\begin{equation}
\label{eq:elbo-general}
\elbo(\mbphi) = \ELL{\mbphi}{\Yobs, \mbX, \mbA}  - \KL{ q_{\mbphi}(\mbA)}{p(\mbA)} ,
\end{equation}
where $\ELL{\mbphi}{\Yobs, \mbX, \mbA} \equiv \E_{q_{\mbphi}(\mbA)}   \log p_{\mbtheta}(\Yobs \g \mbX, \mbA)$ %$\E_{q_{\mbphi}(\mbA)}   \log p_{\mbtheta}(\Yobs \g \mbX, \mbA)$
 is the \gls{ELL}, i.e.~the expectation of the conditional likelihood over the approximate posterior, and  $\KL{ q_{\mbphi}(\mbA)} {p(\mbA)}$ is the \gls{KL} divergence between the approximate posterior and the prior.  
 % When using discrete (Bernoulli) prior and posterior distributions the \gls{KL} term can be determined analytically and the \gls{ELL} is estimated via \gls{MC} samples. When using the binary Concrete relaxations for the prior and the posterior, one requires sampling from the approximate posterior and evaluation of  $\log q_{\mbphi}(A_{ij})$. These can be done straightforwardly and details can be found in the supplement. As described above we can use the score-function estimator or the re-parameterization trick within gradient-based optimization of the \gls{ELBO} when we have discrete distributions or their continuous relaxations, respectively. For completeness, here we give the full expression of 
 The \gls{ELBO} when using the Concrete relaxations under a more numerically stable parameterization 
 (see %\cref{app:concrete-distros}  
 the supplement for details) is given by 
%In our experiments, we consider all four variants of free vs smooth parameterizations and discrete vs relaxed distributions. 
%
\begin{equation}
%\begin{multline}
\elbo(\mbphi)  = \E_{g_{\mbphi,\temposterior}(\mbB)} \left [ \log p_{\mbtheta}(\Yobs \g \mbX, \sigma(\mbB)) 
%\right.  
%\\
%\left.
 - \log \frac{g_{\mbphi,\temposterior}(\mbB)}{f_{\temprior}(\mbB)} \right ],
%\end{multline}
\end{equation}
where
\begin{equation}
%\begin{split}
%\ELL{\mbphi}{\Yobs, \mbX, \mbA} &=  \E_{g_{\mbphi,\temposterior}(\mbB)}  \log p_{\mbtheta}(\Yobs \g \mbX, \sigma(\mbB)) \\
%\KL{ q_{\mbphi}(\mbA)}{p(\mbA)} &= \E_{g_{\mbphi,\temposterior}(\mbB)} \log \frac{g_{\mbphi,\temposterior}(\mbB)}{f_{\temprior}(\mbB)} \\
g_{\mbphi, \temposterior} (B_{ij}) = \Logistic\left(B_{ij} \g \frac{\log \bernrelaxedposterior_{ij}}{\temposterior}, \frac{1}{\temposterior}\right), \quad
  f_{\temprior} (B_{ij}) = \Logistic\left(B_{ij} \g \frac{\log \bernrelaxedprior_{ij}}{\temprior}, \frac{1}{\temprior}\right), 
%\end{split}
\end{equation}
%\end{equation}
 %$g_{\mbphi, \temposterior} (B_{ij}) = \Logistic(B_{ij} \g \frac{\log \bernrelaxedposterior_{ij}}{\temposterior}, \frac{1}{\temposterior})$; $f_{\temprior} (B_{ij}) = \Logistic(B_{ij} \g \frac{\log \bernrelaxedprior_{ij}}{\temprior}, \frac{1}{\temprior})$; 
$\sigmoid(\mbB)$ computes the entrywise logistic sigmoid function over $\mbB$; $\Logistic(B \g \mu, s)$ denotes a Logistic distribution with location $\mu$ and scale $s$  and the distributions $g_{\mbphi,\temposterior}(\mbB)$ and $f_{\temprior}(\mbB)$ factorize over the entries of $\mbB$. The expectation $E_{g_{\mbphi,\temposterior}(\mbB)}$ is estimated using $S$ samples from the re-parameterized posterior, which can be obtained using \cref{eq:sampling-A} below. Estimation of the variational parameters $\mbphi$ is done via gradient-based optimization of the \gls{ELBO} with the gradients obtained by automatic differentiation.  
\subsection{Predictions}
The posterior predictive distribution of the latent labels, given our factorized assumption of the conditional likelihood in \cref{eq:likelihood}, can be obtained as 
\begin{equation}
	p(\Ylatent \g \Yobs, \mbX) = \sum_{\mbA} p_{\mbtheta}(\Ylatent \g \mbX, \mbA) p(\mbA \g \Yobs, \mbX)  % \approx  \sum_{\mbA} p(\Ylatent \g \mbX, \mbA) q_{\mbphi}(\mbA) 
	 \approx \frac{1}{S} \sum_{s=1}^{S} p_{\mbtheta}(\Ylatent | \mbX, \Asample{s}) ,
\end{equation} 
where $\Asample{s}$ is a sample from the posterior $q_{\mbphi}(\mbA)$, $S$ is the total number of samples and $p_{\mbtheta}(\Ylatent \g \mbX, \mbA)$ is the \gls{GCN}-likelihood given in \cref{eq:likelihood}. These samples can be obtained as: 
\begin{equation}
	\label{eq:sampling-A}
	U  \sim \Uniform(0,1), \quad
	 \asample{s}{ij} = \sigmoid\left(\frac{\log \bernrelaxedposterior_{ij} + \log U - \log (1 - U)}{\temposterior}\right) ,
\end{equation}
where, as before,  $\{\bernrelaxedposterior_{ij}\}$ are the estimated parameters of the posterior and $\temposterior$ is the posterior temperature.

\subsection{Computational complexity}
\label{sec:complexity}
We require to compute $\bigO(N^2)$  individual \gls{KL} divergences,  which can be trivially parallelized. 
%In the case of the smooth parameterization, for both the discrete and the relaxed cases, we need to compute the dot-product between the latent representations for each each $A_{ij}$ which is $\bigO(d_z)$ and gradient information must be aggregated for each $\mbz_i, \tilde{\mbz}_i$.  
While for a discrete posterior these individual \gls{KL} terms can be computed exactly (as shown in the supplement),  for the continuous relaxation we need to resort to  \gls{MC} estimation over $S$ samples.  Aggregation over samples can also be parallelized straightforwardly.
%
%\textbf{\gls{ELL} term}: 
Computing the \gls{ELL} using a 2-layer \gls{GCN} as in \cref{eq:gcn-two-layers} requires $\bigO(NDQ + S(NQC + N^2Q + N^2C))$ for the continuous case. However, in the discrete case it only requires doing a forward pass over the standard \gls{GCN}  architecture $S$ times, hence being linear in the number of edges, \ie $\bigO(S |\mathcal{E}| D Q C)$, where $|\mathcal{E}|$ is the expected number of edges sampled from the posterior, assuming sparse-dense matrix multiplication is exploited. In order to reduce the number of parameters and allow for mini-batch training, our approach can be combined with other methods such as cluster-\gls{GCN} \citep{ClusterGCN}. We present an example of this as well as more details of our method's computational complexity in the supplement.
% Further details can be found in the supplement

%In terms of memory, our algorithm requires drawing from the approximate posterior over $\mbA$ explicitly. When using the binary Concrete distribution the leading factor is  $\bigO(N^2)$ as we do not require to keep the samples simultaneously to compute the empirical expectations. 

%\input{model2}
\section{Experiments}
\label{sec:experiments}
In this section we describe the experiments carried out to evaluate our method, which we will refer to as 
%variational Bayesian approach and compare it with several benchmark models. Throughout this section and in the supplement, we will refer to our method as 
\gls{VGCN}\footnote{Code available at \url{https://github.com/ebonilla/VGCN}.}. We will focus on its relaxed version when using the free parameterization with binary Concrete prior and posterior distributions, which we found to be much more stable during training than when using the discrete version and outperformed the low-rank parameterization under  latent-dimensionality constraints (see the supplement for details). 
%
% \subsection{Experimental Set-Up}
We will start by analyzing the %behavior of our algorithm and that of several competing graph neural network baselines 
scenario when no input graph is given and will refer to this  as the \emph{no-graph} case. This is motivated by the fact that in many practical applications graphs are created in an ad hoc basis based on side information or node features \cite{henaff2015deep}. 
% THis is repeated below
% In this case, our prior is constructed as a function of the input features using a \gls{KNNG}, as described in \cref{sub:prior}.
%
Then we present results on the robustness of the algorithm when there is an input graph associated with the given dataset and such a graph is subjected to adversarial perturbations. We refer to this scenario as the \emph{adversarial setting}. We conclude the section with an analysis of the estimated posterior distribution and by showcasing the performance of our method when using the ground truth (unperturbed) graphs.   
% The corrupted graphs are generated using the recently proposed framework of \cite{pmlr-v97-bojchevski19a}, which has been shown to be much more effective than uniformly-at-random corruption and for which graph neural network algorithms have been shown to be highly vulnerable. We consider perturbations that add or remove links to the given graph.  

%\subsection{Datasets}
\subsection{Experimental set-up}
\parhead{Datasets.} We use two citation-network datasets (\cora and \citeseer) and an additional graph of political blogs (\polblogs). In the citation networks the nodes represent documents and their links refer to citations between documents. We construct an undirected graph based on these citation links so as to
obtain a binary adjacency matrix. Each document is characterized by a set of features, which are either \gls{BOW} or \gls{TFIDF}
for \cora and \citeseer, respectively.  
The class for each document is given by their subject and our goal is to predict this for a subset of unlabeled documents.
The \polblogs dataset is a network of political blogs first introduced in \cite{adamic2005political}. For our experiments, we use the dataset as published in \cite{pmlr-v97-bojchevski19a}. Nodes represent blogs and two nodes are connected if a blog links to another from anywhere on the landing page. Each node is labeled as one of two classes based on the blog's political affiliation (conservative vs liberal). No node features are available and so we use an identity matrix as is common practice. 
Details of these datasets are given in %\cref{sec:datasets} 
the supplement.

%\subsection{Training Details and Baselines}
\parhead{Training details.}
For the citation networks we used a similar setting to that in \cite{yang2016revisiting} where 20 labeled examples per class are used for training, 1,000 examples are used for testing and the rest are used as unlabeled data. We are very much aware of the potential difficulties when using fixed training datasets for evaluation in machine learning, in general, and in particular in graph neural networks \citep{shchur2018pitfalls}. However, we believe our experiments introduce enough additional randomization so that the results can be considered as reliable. Firstly, we adopt the set-up in the recently proposed work of  \cite{lds-2019}, where the original splits are augmented so as to include 50\% of the validation set. Furthermore, in the no-graph case,  we generate 10 different versions of each dataset and also construct \glspl{KNNG} using $k=\{10, 20\}$ and the cosine and Minkowski distances. In the adversarial setting,  also using the augmented datasets, we explore 7 different noise-level aggregations and replicate each experiment 10 times. To alleviate the phenomenon % of latent variable collapse, \todo{Address complaint re posterior collapse.} 
of the variational posterior collapsing to the prior and thereby failing to learn latent representations that explain the data~\cite{bowman2016generating, dieng2019avoiding}, we dampen the effect of the KL regularization term in the ELBO by scaling via a parameter $\beta < 1$~\cite{higgins2017beta, alemi2018fixing}. This 
%hyper-parameter (along with others such as $\smoothfactorone$ and temperatures) 
and other hyper-parameters were tuned using cross-validation, see the supplement for details.

\parhead{Baselines and performance metrics.}
We compare against standard \gls{GCN} \cite{kipf2016semi}, \acrlong{GRAPHSAGE}\footnote{\gls{GRAPHSAGE} has been shortened to SAGE in the figures.} \citep[\acrshort{GRAPHSAGE},][]{hamilton2017inductive} and \acrlongpl{GAT} \citep[\acrshortpl{GAT},][]{velickovic2018graph}, which are all competitive graph neural network algorithms that assume a noise-free input graph is given. Additionally, we  benchmark our algorithm against the \gls{LDS} framework of \cite{lds-2019} which, like ours, attempts to learn a graph generative model for \glspl{GCN} (see \cref{sub:related_work} for more details). We consider other Bayesian approaches to \gls{GCN}, in particular, the work of \cite{zhang2018bayesian} which herein we refer  to as \gls{EGCN} and the \gls{GGP} approach of \cite{zhang2018bayesian}. Finally, in the adversarial case we compare against \gls{RGCN} of \cite{robustgcn2019} that extends \gls{GCN} for robustness to adversarial attacks. Overall, we believe this set of 7 benchmarks provides state-of-the-art competing algorithms that show a realistic and up-to-date evaluation of the benefits of our approach.
In terms of performance metrics, throughout our experiments we use the test accuracy as given by the proportion of correctly classified test examples and use the validation accuracy for model selection on all methods.  More details of the experimental set-up % including hyper-parameter setting 
are given in the supplement.
%
%\paragraph{Performance measures}
% using validation accuracy for model selection
%
%
%
%Hyper-parameter tuning \ebnote{complete here or in the appendix}
%
%
%\subsection{Results}
%
\subsection{Results in the no-graph case}
\begin{figure}[t]
%  \begin{minipage}[t]{\textwidth}
	\centering
	\includegraphics[width=0.4\textwidth]{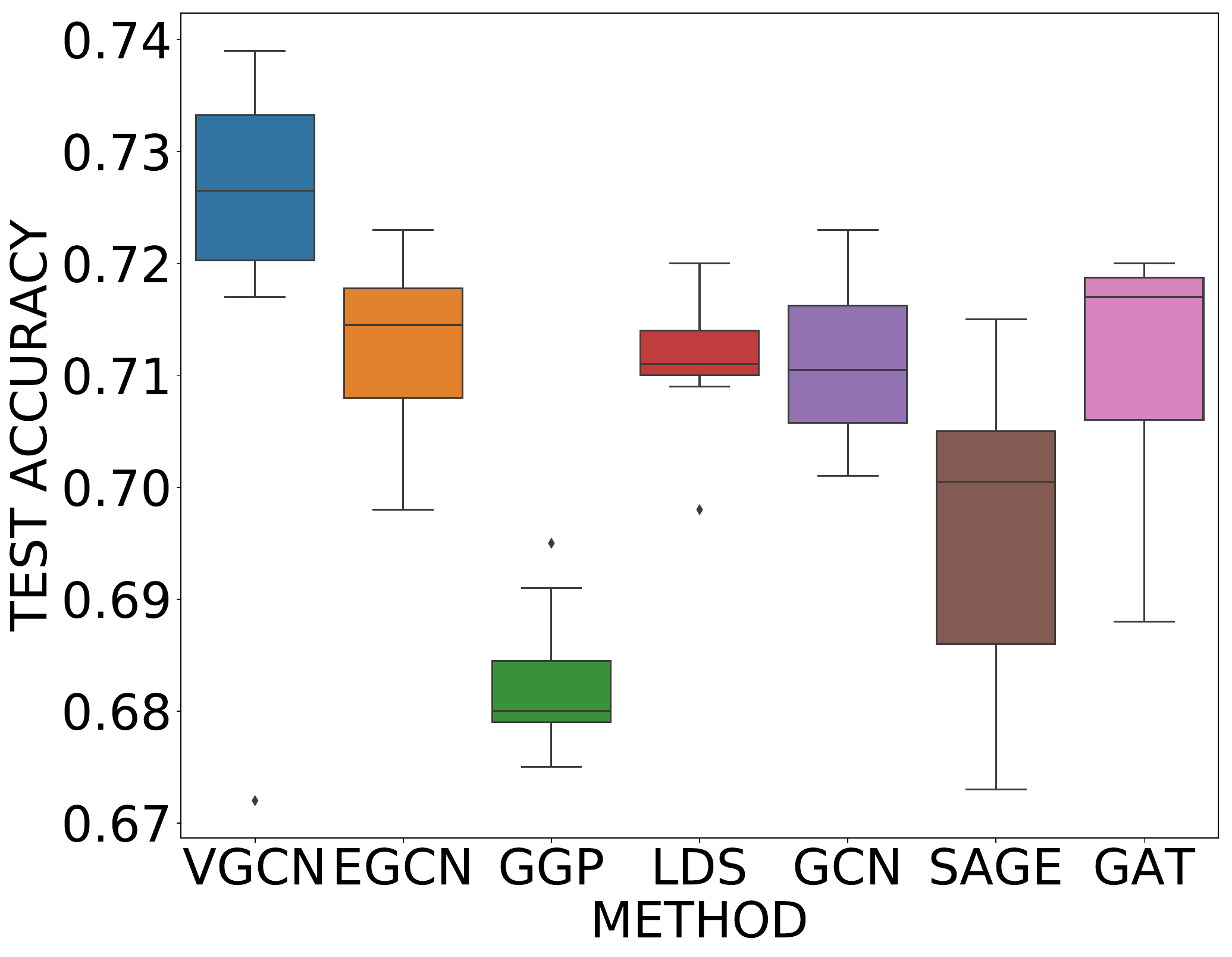}
	~
	\includegraphics[width=0.4\textwidth]{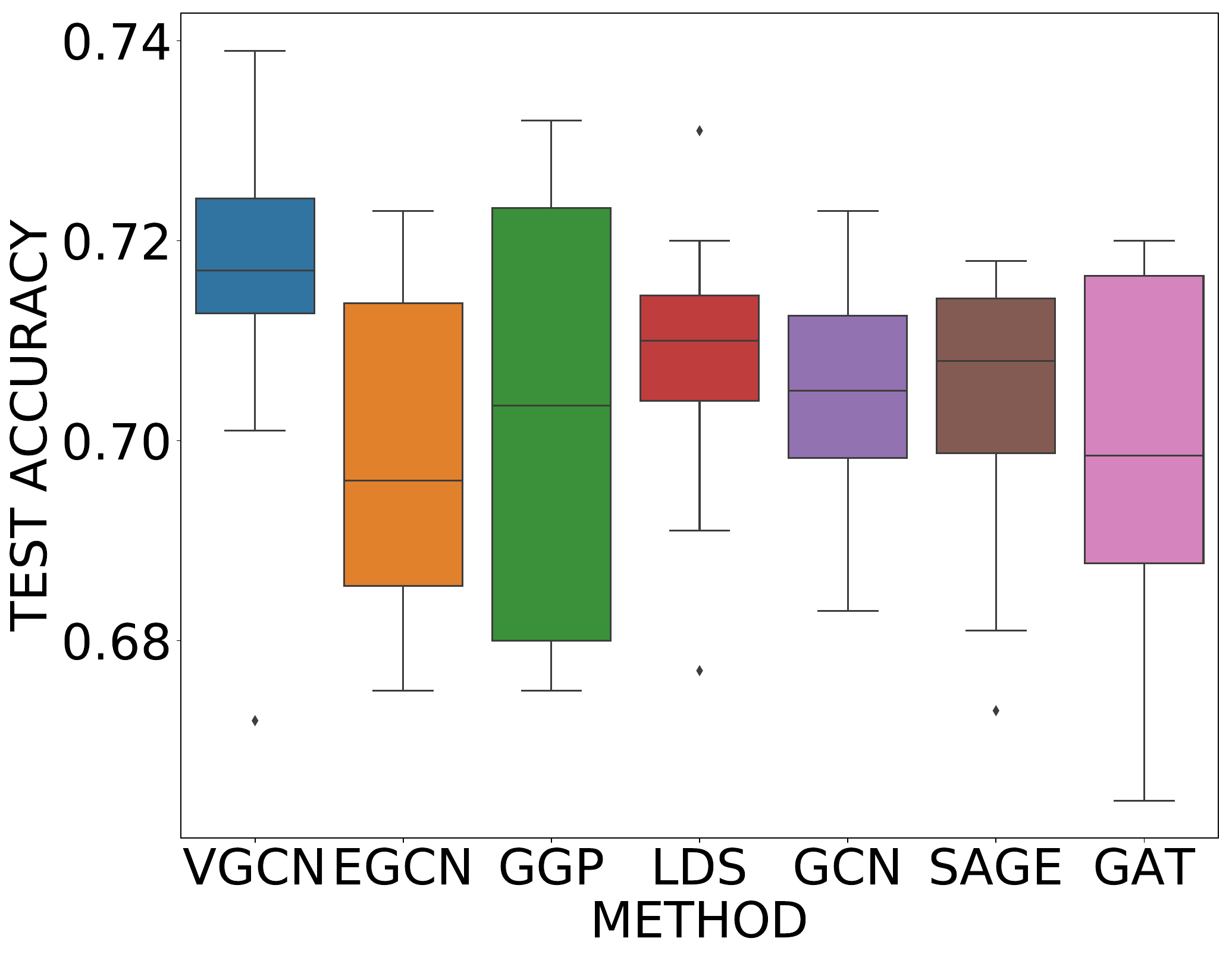}
	\caption{Test accuracy for the no-graph case on \citeseer (left) and \cora (right) across ten replications for our method (\gls{VGCN}) and competing algorithms. The first three methods are Bayesian, while the others are not.
	\label{fig:knngraph-relaxed-accuracy} }
%	\end{minipage}
%	\hfill
% 	\begin{minipage}[t]{\textwidth} 
% 	\centering
% 	\includegraphics[width=0.45\textwidth]{Citeseer_attacked_graphs_accuracy_final.pdf}
% 	\includegraphics[width=0.45\textwidth]{Cora_attacked_graphs_accuracy_final.pdf}
% 	\caption{Test accuracy for the adversarial setting on attributed graphs \citeseer (left) and  \cora (right) when removing (negative values) or adding (positive values) edges. Our method is denoted by\gls{VGCN}. 
% 	\label{fig:attackedgraphs-relaxed-accuracy} }
%     \end{minipage}
\end{figure}

%parhead{No-graph case.}
Here we present the performance of our proposed method (\gls{VGCN}) and the baselines' when we do not use the graph that is associated with the corresponding citations network. In this case % as described above, 
we build a \gls{KNNG} and use it to construct a prior distribution (as described in \cref{sub:prior}) for the Bayesian methods (\gls{VGCN}, \gls{EGCN}, \gls{GGP}) or directly for the non-Bayesian algorithms (\gls{LDS}, \gls{GCN}, \acrshort{GRAPHSAGE}, \acrshort{GAT}). We see that on \citeseer (left of \cref{fig:knngraph-relaxed-accuracy}) our algorithm outperforms both Bayesian and non-Bayesian methods, with the Bayesian \gls{GGP} approach of \cite{zhang2018bayesian} providing the lowest test accuracy. Although there is no clear distinction of Bayesian vs non-Bayesian methods, these results are not incredibly surprising as the non-Bayesian methods optimize the cross-entropy error directly. 

The benefits of our approach are less pronounced on \cora as seen on the right of \cref{fig:knngraph-relaxed-accuracy}, while we still see a marginal improvement over the other baselines. We attribute these differences between the datasets to the types of features used, \gls{BOW} vs \gls{TFIDF}. Overall we can conclude that our method does manage to discover new graph parameterizations that improve performance even over methods that were specifically designed to do so, such as the \gls{LDS} algorithm of \cite{lds-2019}. We also note that, as reported in \cite{lds-2019},  a dense \gls{GCN} that does not use the graph (corresponding to  a multi-layer Perceptron) achieves test accuracies of 58.4\% and 59.1\% on \citeseer and \cora respectively.  

\subsection{Results in the adversarial setting}
%\parhead{Adversarial Setting.}
We consider the robustness of \gls{VGCN} and the baselines in the adversarial setting where the ground truth graph has been corrupted via the removal or addition of edges. Specifically, we consider the graph poisoning setting as outlined in~\cite{pmlr-v97-bojchevski19a} where the graph is poisoned before model training. We limit our study on general attacks where the attacker uses an unsupervised approach to perturb network structure and is not targeted to a classification task. In~\cite{pmlr-v97-bojchevski19a}, it was shown that poisoning the network structure reduces performance on downstream tasks and transfers across to graph convolutional methods. These graph poisoning attacks are the easiest for an attacker to deploy in practice.

We applied graph poisoning on all 3 of our datasets removing 2,000, 1,000, and 500 edges (denoted using negative values in the figures) and also adding 500, 1000, 2,000, and 5,000 edges. We generated 10 attacked graphs for each of these settings. We note that removing 5,000 edges is not actually possible on \citeseer hence we do not consider this setting. We show results for the extreme cases -2,000, 2,000, and 5,000 here and for all settings in the supplement. 

\textit{Citation Networks}: 
\Cref{fig:attackedgraphs-relaxed-accuracy} (left and middle) shows results for attacked {graphs with node features}, namely the two citation networks. We see that all methods perform well in the -2000 setting where the graphs are missing edges but all remaining edges are uncorrupted. The Bayesian methods \gls{VGCN} and \gls{EGCN} have an advantage over the others on \citeseer but  the algorithms' performance is more leveled up on \cora. We note that performance for all methods degrades when false edges are added to the graphs. \gls{VGCN} is the most robust especially in the extreme case of adding 5000 edges, effectively doubling the total number of edges in the graphs for both datasets. \gls{VGCN} outperforms \gls{RGCN}, which was specifically designed for these types of problems,  especially in the cases of adding edges. Lastly, we note that variance for \gls{VGCN} is lower across all graphs and datasets.

\textit{Featureless Graphs}:
Finally, we evaluate the  performance of all methods on a {graph without node features}.  %\Cref{fig:attackedgraphs-polblogs-relaxed-accuracy} 
\Cref{fig:attackedgraphs-relaxed-accuracy} (right) 
shows results for the \polblogs network where node features are not available. All methods perform competitively across all attacked graph settings. Surprisingly, methods such as \gls{GRAPHSAGE} and \gls{GAT} show superior performance for the -2000 setting but have very high variance at the 5000 regime. \gls{RGCN} performs best when removing edges but does poorly when adding edges due to its heavy reliance on the similarity of node features. 
\begin{figure}
 	\centering
 	\includegraphics[width=0.32\textwidth]{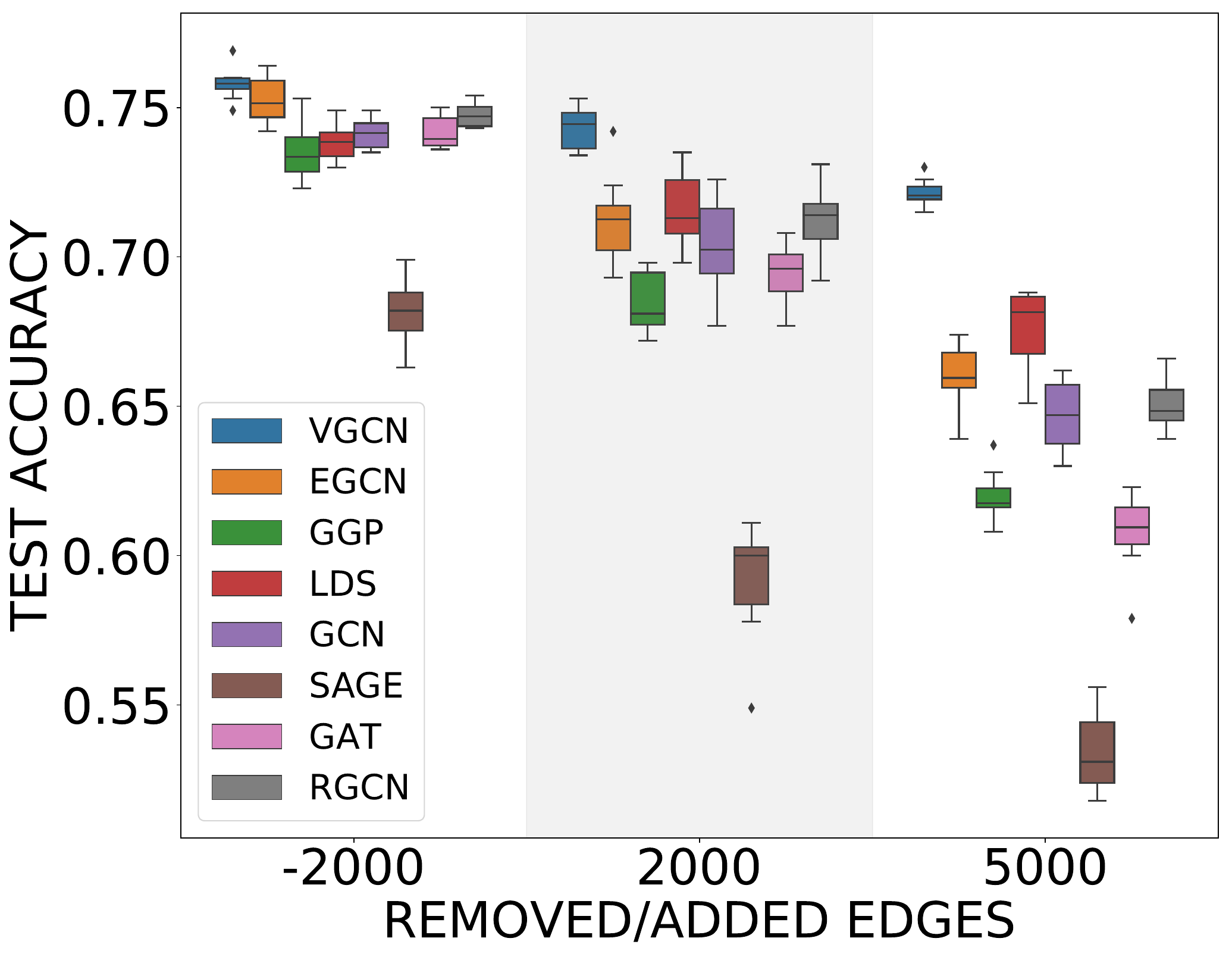}
 	\includegraphics[width=0.32\textwidth]{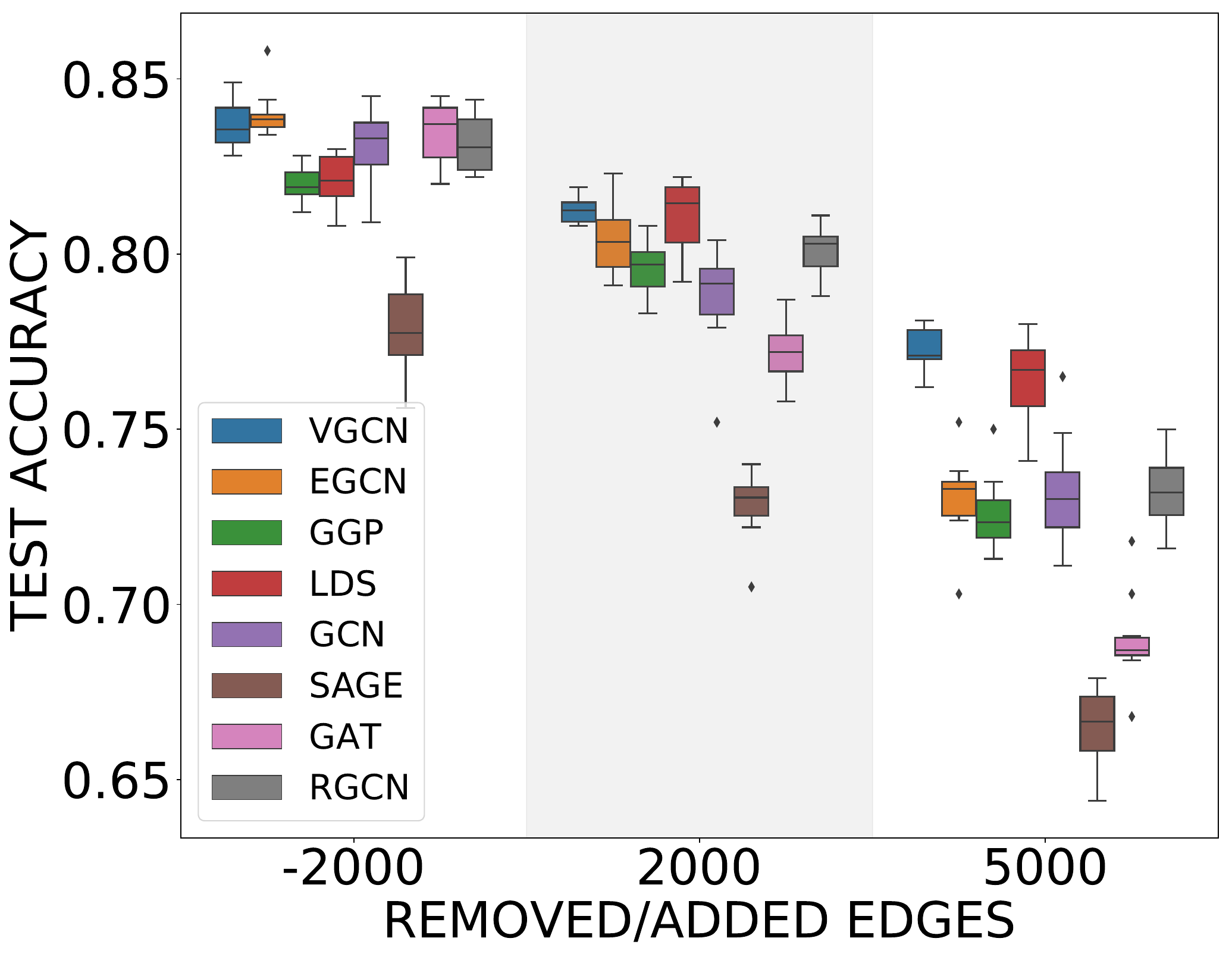}
 	\includegraphics[width=0.32\textwidth]{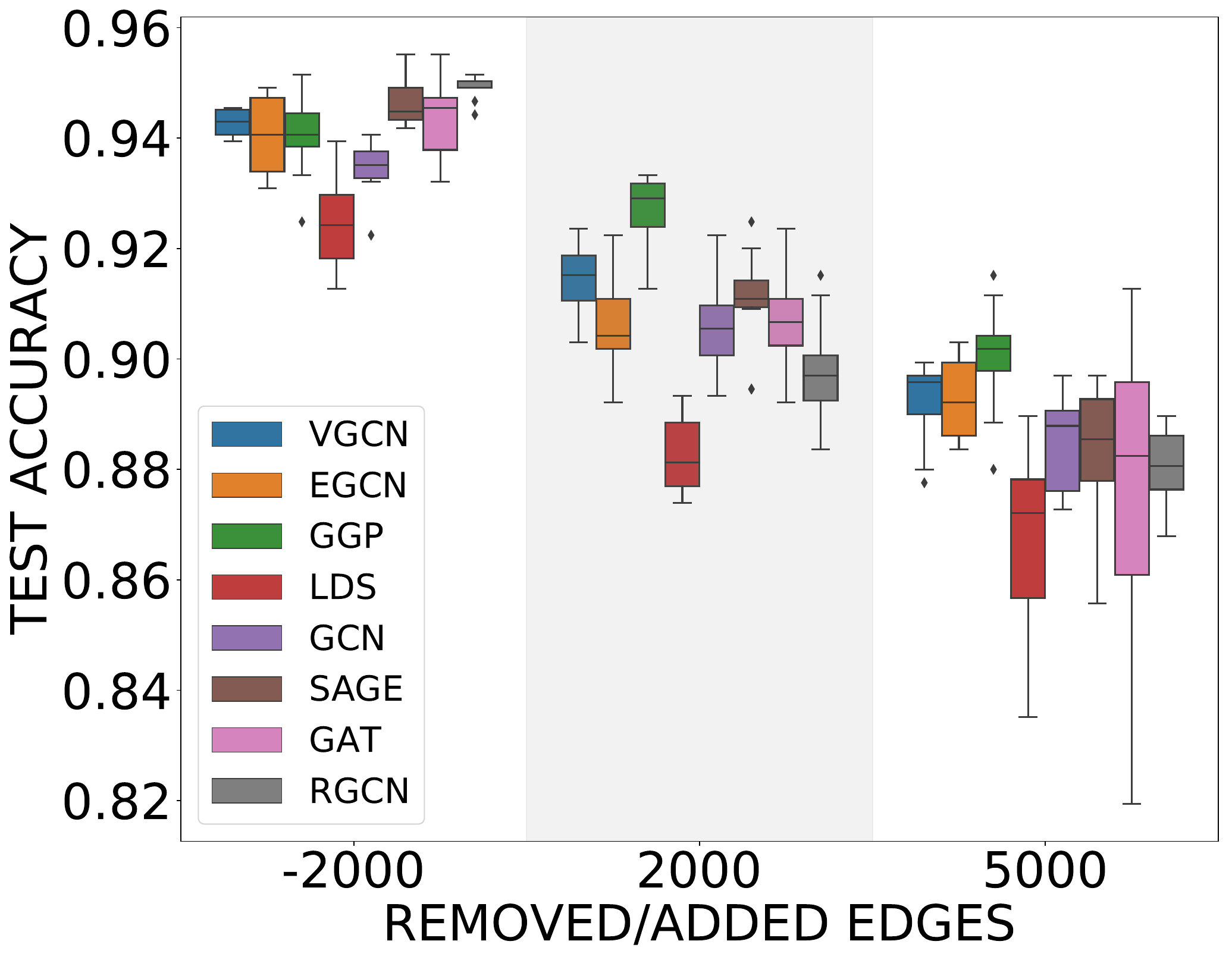}
 	\caption{Test accuracy for the adversarial setting on attributed graphs \citeseer (left) and  \cora (middle)
 	and featureless graph \polblogs (right) 
 	when removing (negative values) or adding (positive values) edges. Our method is denoted by \gls{VGCN}. 
 	\label{fig:attackedgraphs-relaxed-accuracy} 
 	}
\end{figure}

% \begin{wrapfigure}{r}{0.32\textwidth}
% 	\includegraphics[width=0.32\textwidth]{PolBlogs_attacked_graphs_accuracy_final.pdf}
% 	\caption{Results for the adversarial setting on \polblogs.} 
% 	% (description as in \cref{fig:attackedgraphs-relaxed-accuracy})}
% 	\label{fig:attackedgraphs-polblogs-relaxed-accuracy}
% \end{wrapfigure}
%
While the Bayesian methods, \gls{VGCN}, \gls{EGCN}, and \gls{GGP} are the most robust on this dataset,  \gls{LDS} is the worst performer across all graphs. One possible explanation for this is  that the lack of node features negatively affects methods that optimize the graph structure as there may not be enough information in the training data and the graph structure alone to optimize the models with higher learning capacities, i.e., more parameters. Furthermore, we note that since the ground truth graph has approximately 16,000 edges, the 5000 fake edges are only an additional ~30\% as compared to  nearly doubling of the number of edges for the citation networks. For higher levels of noise, Bayesian methods might demonstrate higher levels of robustness. % We leave further study with increased levels of corruption for future work. 
%
%\subsection{Results on Ground Truth Graphs Case and Posterior Analysis}
\subsection{Performance on ground-truth graphs and qualitative analysis}
\begin{figure}[t]
    \centering
    \includegraphics[width=0.32\textwidth]{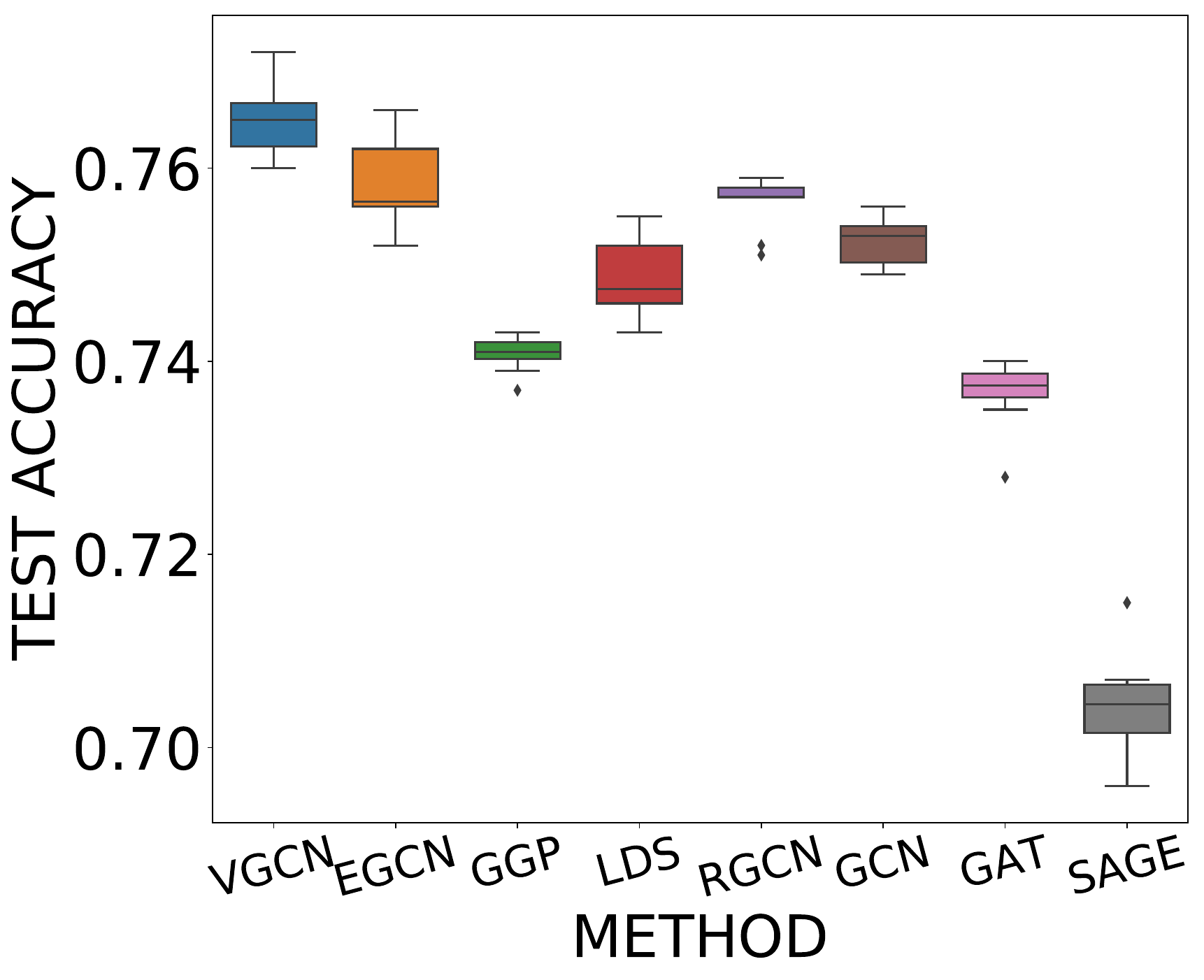}
    \includegraphics[width=0.32\textwidth]{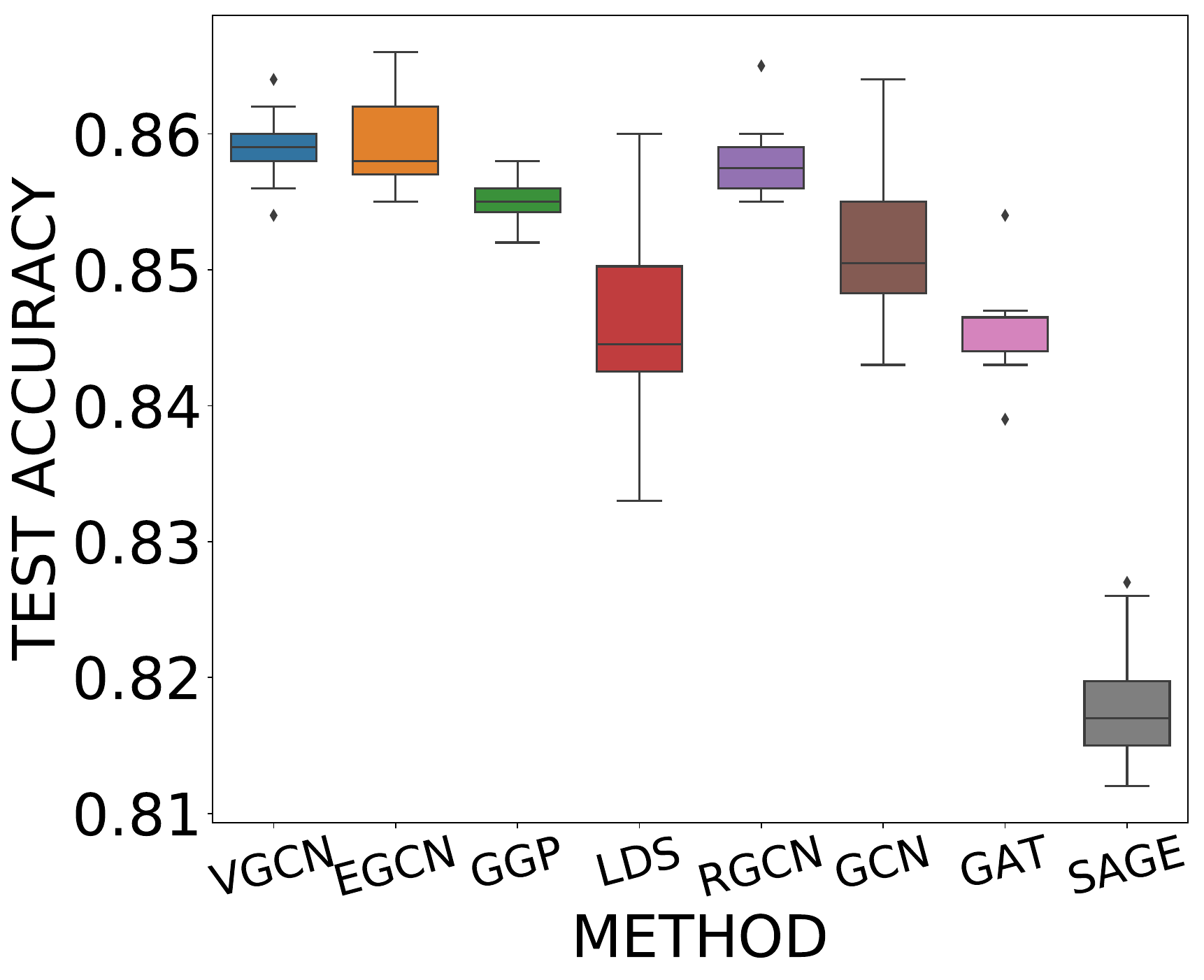}
    \includegraphics[width=0.32\textwidth]{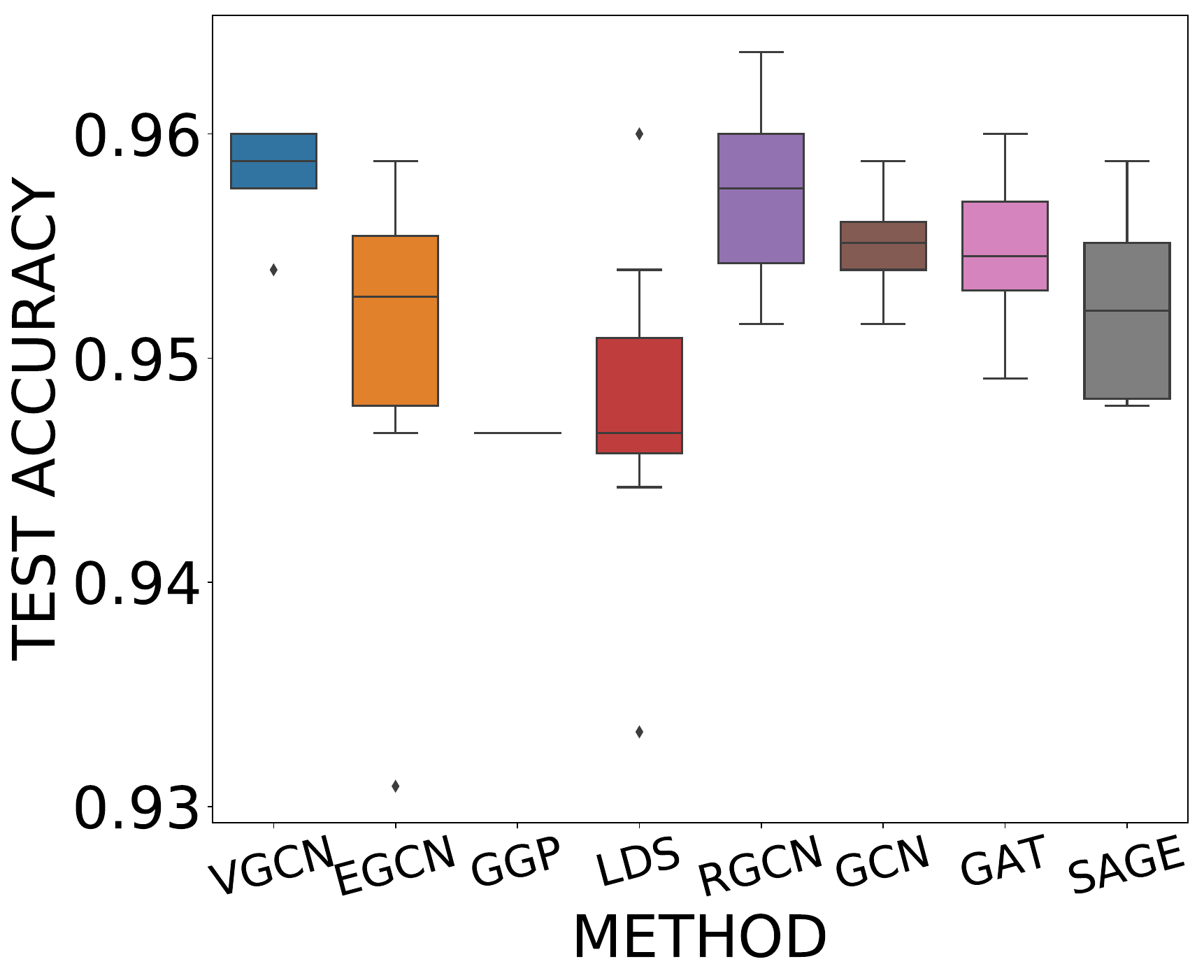}
    \hfill
    \caption{\textit{From left to right}: Test accuracy for ground-truth graphs  on  \citeseer, \cora and \polblogs across ten replications for our method (\gls{VGCN}) and competing algorithms.
    %\textit{Right}: The limit posterior probabilities of a link being turned on computed as the zero-temperature limit of the final variational posterior distributions over each adjacency entry (see text for details), using the \citeseer dataset under the no-graph scenario.
    \label{fig:knowngraph-relaxed-accuracy}
    }
\end{figure}

\parhead{Performance on ground-truth graphs.}
%Here we showcase the performance of\gls{VGCN} for the ground-truth graphs. In this case, we demonstrate that \gls{VGCN} performs as well as and in some cases better than the baselines. In \cref{fig:knowngraph-relaxed-accuracy} we show the results for the 3 datasets under consideration. 
% Finally,
\Cref{fig:knowngraph-relaxed-accuracy}  shows the results for the ground-truth (unperturbed) graphs, where we see that \gls{VGCN} can provide state-of-the-art performance, 
%We see that \gls{VGCN} outperforms all other methods on \citeseer. \gls{GCN} and \gls{RGCN} are the next two best performing methods whereas \gls{GRAPHSAGE} performs worse on \citeseer and \cora but outperforms \gls{LDS} on \polblogs. This indicates that \gls{VGCN} can improve performance for semi-supervised node classification over existing methods by explicitly modeling the quality of the available edges.
indicating that it can find better configurations even for those graphs that are believed to be most beneficial for the semi-supervised node classification task.
\label{sec:qualitative_analysis}
%\subsection{Posterior analysis}

\begin{figure}[t]
\centering
\includegraphics[width=0.32\textwidth]{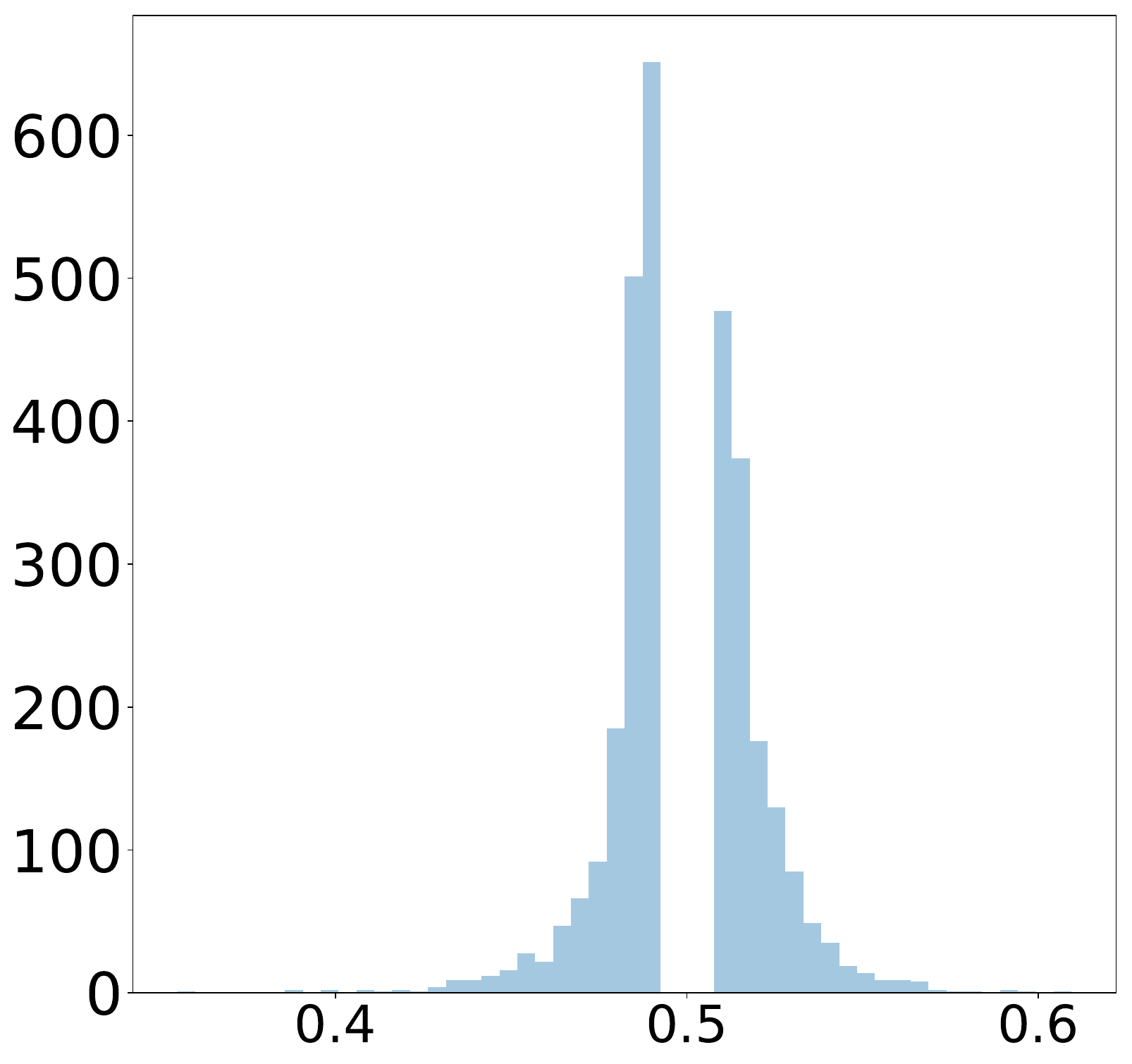}
\includegraphics[width=0.32\textwidth]{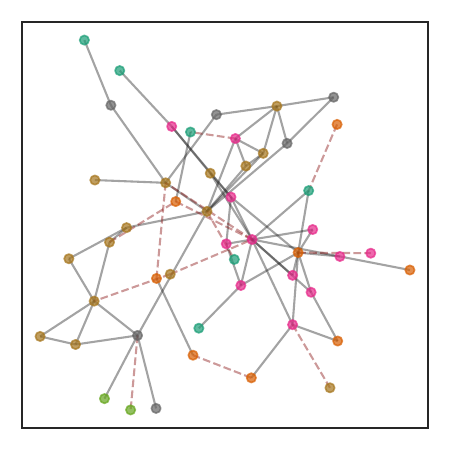}
\includegraphics[width=0.32\textwidth]{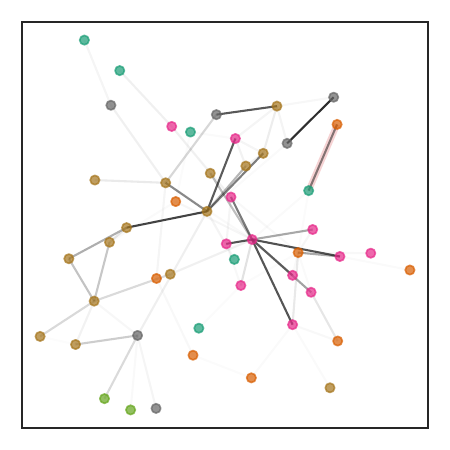}
\caption{\emph{Left:} Histogram of limit posterior probabilities of a link being turned on computed as the zero-temperature limit of the final variational posterior distributions over each adjacency entry (see text for details), using the \citeseer dataset under the no-graph scenario.
\emph{Middle:} 
A community in the original graph from the adversarial \citeseer experiment with node labels distinguished by colors and added edges denoted by red dashes. 
\emph{Right:}  Learned graph with edge opacity proportional to limit posterior probabilities. Added edges with probability greater than \nicefrac{1}{2} are highlighted red. 
\label{fig:qualitative-analysis}
}
\end{figure}

\parhead{Limit posterior probabilities.}
\label{sec:posterior-analysis}
Here we analyze what our model has learned and how different the resulting posterior is with respect to the prior and its initialization during optimization.  
% For this purpose we recall that our main results are based on the free parameterization of the posterior under the continuous relaxation as given by the binary Concrete distribution with parameters $\lambda_{ij}$ and temperature $\tau$. 
% 
Interpreting Concrete distributions can be cumbersome so instead we use their zero-temperature property as presented in \cite{maddison2016concrete}. More specifically, in the zero-temperature limit, one can obtain the corresponding Bernoulli parameters of their discrete counterpart using
%\begin{equation}
%	\label{eq:prob-limi}
$
	p(\text{lim}_{\tau \rightarrow 0}  A_{ij} = 1) = {\lambda_{ij}}/{( \lambda_{ij} + 1)}
$.
%\end{equation}
%
%\begin{figure}[]
%	\centering
%	\includegraphics[width=0.23\textwidth]{sections/results_posterior_knn-graph.pdf}
%	\caption{The limit posterior probabilities of a link being turn on computed as the zero-temperature limit of the final variational posterior distributions over each adjacency entry (see text for details), using the \citeseer dataset under the no-graph scenario. Only showing those probabilities that changed significantly from the prior, which has a peak at 0.5 and which was used to initialize the posterior.   
%		\label{fig:posterior-knn} }
%\end{figure}
% Therefore, we have computed the corresponding Bernoulli probabilities of our final variational posterior over the adjacency matrix using Equation \cref{eq:prob-limi}. 
We refer to these probabilities as the \emph{limit posterior probabilities}.  
\Cref{fig:qualitative-analysis} (left) shows a histogram of these limit probabilities on a typical run of our model in the no-graph case for the \citeseer dataset.
%, when using a smoothing factor $\smoothfactorone = 0.5$, which was a typical selection of our algorithm after  cross-validation.  
% Recall that, in the no-graph case,  our prior is built using a \gls{KNNG} graph and smoothed out accordingly and that the posterior was initialized to the prior.  
As the majority of the links are close to zero, \cref{fig:qualitative-analysis} (left) only shows those probabilities that changed significantly with respect to the prior (defined as their absolute difference being greater than $0.02$). Hence, we see that in effect our algorithm manages to both decrease and increase these initial probabilities, providing evidence that it has the capacity to turn links on/off in the original graph. More precisely, the number of links with a significant change in their probabilities of being turned on was 3046. We believe this is indeed a significant amount, considering that the \acrlong{KNNG} had around 33,000 links and that the `true' \citeseer graph has around 4600 edges. 

\parhead{Learned graphs.}
We illustrate the types of graphs learned by our approach by using an example from the adversarial setting experiment on \citeseer when adding edges, where our approach significantly outperformed the competing baselines. 
Since showing the entire graph would be unintelligible, we enumerate communities (subgraphs that are internally densely connected) that contain a good balance of node labels.
In particular, we use label propagation \cite{zhu-2002-label-propagation} to 
detect the largest communities and draw a graph for each. 
We show an example subgraph in \cref{fig:qualitative-analysis} (middle and right). On the middle we denote the edges from the original graph in solid lines and the added edges in dashed red lines. On the right is the corresponding complete graph with edge opacity proportional to their limit posterior probabilities. 
Generally speaking, the posterior probabilities of the original edges can be expected to remain largely the same or in some cases, either amplified or attenuated to improve downstream classification accuracy. More interestingly, we see that, with few exceptions (highlighted in red), the posterior probabilities of the added edges are attenuated. We show more example subgraphs in the supplement.

% To qualitatively analyze our approach, we examine densely-connected subgraphs 
% of the \textsc{citeseer} graph used in the experiment shown in (i.e.~adding edges). 

% Preliminary results in the given figure (node colors indicate node labels; left: original corrupted graph with red dashed lines indicating added edges; right: learned graph shown with edge opacities proportional to posterior probabilities; best seen zoomed in on screen) indicate that, with a few exceptions, the posterior probabilities of the \emph{added} edges are indeed attenuated.

\section{Conclusion \& discussion}
\label{sec:conclusion}
%We have presented a variational inference approach to  Bayesian graph convolutional networks by building upon state-of-the-art neural network architectures for graph-based semi-supervised classification. We have shown that \gls{GCN} machinery can be beneficial even in the absence of an input graph. Furthermore, we have made \gls{GCN} more robust to adversarial attacks.   Our method deals with these problems in a natural way by considering prior distributions over the adjacency matrix along with a \gls{GCN} likelihood in a joint probabilistic model. Our inference algorithm for posterior estimation relies on a prior and posterior relaxations and can outperform state-of-the-art graph neural network algorithms.  An immediate direction for future work is to address the scalability of our method, as the relaxed posterior version cannot exploit sparse computations. %A related avenue is to improve upon the discrete posterior version of our algorithm as it does exploit sparse operations but uses high-variance gradient estimates.
%Finally, it is important to emphasize that our method can be extended to use other graph neural network architectures as long as the graph is represented by an adjacency matrix and that parameter estimation in the original method can be combined easily with variational inference.  %We intend on exploring this avenue in the future. 
We have presented \acrfullpl{VGCN} for semi-supervised classification, a method that generalizes the capabilities of \glspl{GCN} by making them applicable in the absence of graph data and more robust to adversarial attacks.  \gls{VGCN} considers prior distributions over the graph along with a \gls{GCN} likelihood in a joint probabilistic model and infers a graph posterior exploiting Concrete distributions. We have showcased the performance of our method on the above problems and using the ground-truth graphs.
% It is important to emphasize that 
Our method can be extended beyond \glspl{GCN} to use other graph neural network architectures as long as the graph is represented by an adjacency matrix and that parameter estimation in the original method can be combined easily with variational inference.  %We intend on exploring this avenue in the future. 

\section*{Broader Impact}
There exist numerous useful applications for graph-\gls{CNN}s including e-commerce product recommendations \cite{Wang2020POI}, online social network recommendations \cite{Rex2018Pinsage}, drug discovery \cite{Junying2017, Nguyen684662, Gilmer2017}, computational pharmacology \cite{Zitnik2018}, disease understanding \cite{HENA2019}, bioinformatics \cite{fout2017protein}, finance \cite{Xiao2019}, anti-money laundering \cite{Weber2018}, online hate speech classification \cite{Ribeiro2018}, and understanding online fake news propagation \cite{Monti2019}. 

Of the above applications, several can benefit from \gls{VGCN} over existing graph-\gls{CNN}s. Any application where there is incentive for bad actors to poison the data in order to (a) hide their activities, e.g., online hate speech, (b) control the activities of others, e.g., trick consumers into purchasing unreliable or expensive products, or (c) promote misinformation, e.g., fake news, stand to benefit from \gls{VGCN}'s ability to deal with adversarial attacks. 

Furthermore, \gls{VGCN} is directly applicable to non-graph domains via the ad hoc construction of graphs hence the benefits of graph-\gls{CNN}s can be brought into domains where data is not inherently graph-structured. As we have discussed earlier, such ad hoc graph creation results in noisy graphs requiring the use of principled structure modeling for training useful predictive models. 

Just like any machine learning algorithm can be used for good it can also be used for harm. Graph-\gls{CNN}s and our method \gls{VGCN} are not immune to such misuse. For example, understanding how \gls{VGCN} deals with adversarial attacks can be used by an adversarial agent to create more robust attacks and subvert attempts at detection. Currently, we have no solution for such a general problem but we understand that this needs to be addressed in future work.

% Finally, \gls{VGCN} in its current form does not deal with bias in the data directly. As a result, for some applications \gls{VGCN} predictions may place some individuals in a disadvantaged position. As a hypothetical example, let us consider the case of credit investigation and assignment based on an individual's spatiotemporal activities modeled with a \gls{GCN} \cite{Xiao2019}. In such a case, a person from a minority group might be deemed unworthy of credit due to elements out of their control such as their home address and daily commute. Such an individual might attempt to improve their credit position by employing a poisoning attack such as visiting locations associated with high credit worthiness. Extending and employing a method such as \gls{VGCN} to detect and remove such poisoning data would essentially be re-enforcing existing biases further disadvantaging such individuals. % As it stands, we have no way in \gls{VGCN} for dealing with such situations.

Overall, we believe that our work is beneficial to society because of the many important applications that stand to benefit from \gls{VGCN}'s ability to handle noise in the graph structure.

\begin{ack}
We thank Harrison Nguyen for his contribution to  an earlier version of this paper presented at NeurIPS 2019's Graph Representation Learning (GRL) workshop. 
% EVB was partially funded by DST ...
LT is supported by an Australian Government Research Training Program (RTP) 
Scholarship and a CSIRO Data61 Postgraduate Scholarship.
This work was conducted in partnership with the Defence Science and Technology Group, through the Next Generation Technologies Program.
% EVB thanks the Defence Science and Technology Group for its support via the Next Generation Technologies Program.
\end{ack}

%\documentclass{article}
%\input{preamble_neurips.tex}
% \AtBeginDocument{
%   \newgeometry{
%     textheight=9in,
%     textwidth=7in,
%     top=1in,
%     headheight=12pt,
%     headsep=25pt,
%     footskip=30pt
%   }
% }
%\input{preamble.tex}
%\input{acronyms.tex}
%\input{math.tex}
%\input{vgcn_math.tex}

%\usepackage{xr}
%\myexternaldocument{main_vgcn}

\numberwithin{equation}{section}
%\begin{document}
% \onecolumn

%\input{supp_title}
\appendix
\section{Variational distributions}
In this section we provide more details about the choices of  variatiational distributions over the graph structure. 
\subsection{Variational distribution: free vs smooth parameterizations}
Similarly to the prior definition, our approximate posterior is of the form
\begin{equation}
%\begin{split}
q_{\mbphi}(\mbA) = \prod_{ij} q_\mbphi(A_{ij}),  \text{ with } \quad
 q_{\mbphi}(A_{ij}) = \Bern(A_{ij} \g \berndiscreteposterior_{ij}),  \berndiscreteposterior_{ij}>0,
%\end{split}
\end{equation}
where, henceforth, we use $\mbphi$ to denote all the parameters of the variational posterior. In the case where $\berndiscreteposterior_{ij}$ are free parameters then $\mbphi = \{\berndiscreteposterior_{ij}\}$. We refer to this approach as the \emph{free} parameterization  
We have found experimentally that such a parameterization can make optimization of the \gls{ELBO} wrt $\mbphi$ extremely difficult. Thus, one is forced to either use alternative representations of the posterior, or continuous relaxations of the discrete prior and posterior distributions (see \cref{sec:vi-discrete-relaxed}). Intuitively,  conditional independence in the posterior is  a strong assumption and small changes in $\berndiscreteposterior_{ij}$ will compete with each other to explain the data. Consequently, any continuous optimization algorithm will find it very challenging to find a good direction in this non-smooth combinatorial space.  Therefore, as an alternative, it is sensible to adopt a smooth parameterization:
\begin{equation}
\label{eq:smooth-param}
	\berndiscreteposterior_{ij} = \sigmoid(\mbz_i^T  \tilde{\mbz}_j + b_i + b_j  + s) ,  \mbz_{i}, \tilde{\mbz}_{j} \in \bbR^{d_z},  \{b_{i}, s \in \bbR \},  
\end{equation}
 $i,j=1, \ldots, N$, where $\sigmoid(x) \equiv (1 + \exp(-x))^{-1}$ is the logistic sigmoid function and $d_z \leq D$ is the dimensionality of the parameters $\mbz, \tilde{\mbz}$.  
 As we see, the same representation is shared across the columns and rows of $\mbA$'s Bernoulli parameters, which addresses the combinatorial nature of the optimization landscape of the free parameterization. We note that this parameterization is referred to in the matrix-factorization and link-prediction literature as low-rank \cite{mnih2008probabilistic,menon2011link} or dot-product \cite{kipf2016variational}. In this case the variational parameters are $\mbphi=\{ \{\mbz_i, \tilde{\mbz}_i, b_i\}, s \}$. 
\subsection{Variational distribution: discrete vs relaxed}
\label{sec:vi-discrete-relaxed}
We have defined above a  variational distribution which naturally models the discrete nature of the adjacency matrix $\mbA$. Our goal is to estimate the parameters $\mbphi$ of the posterior $q_{\mbphi}(\mbA)$ via maximization of the \gls{ELBO}. For this purpose we can use the so-called {score function} method \cite{ranganath2014black}, which provides an unbiased estimator of  the gradient of an expectation of a function using \gls{MC} samples. However, it is now widely accepted that, because of its generality, the score function estimator can suffer from high variance \cite{ranganath2016hierarchical}. 
 
Therefore, as an alternative to the score function estimator, we can use the so-called re-parameterization trick \cite{kingma2013auto,rezende2014stochastic}, which generally exhibits lower variance. Unfortunately, the re-parameterization trick is not applicable to discrete distributions so we need to resort to continuous relaxations. In this work we use Concrete distributions as proposed by  \citet{jang2016categorical,maddison2016concrete}. In particular, we  denote our binary Concrete posterior distribution with location parameters $\bernrelaxedposterior_{ij} > 0 $ and temperature $\temposterior > 0$ as $q_{\mbphi}(A_{ij}) = \BinConcrete(A_{ij} \g \bernrelaxedposterior_{ij}, \temposterior)$. Analogously, as discussed in \citet{maddison2016concrete}, in order to maintain a lower bound during variational inference we also relax our prior so that $p(A_{ij}) = \BinConcrete(A_{ij} \g \bernrelaxedprior_{ij}, \temprior)$. 
 In this case the variational parameters are the parameters of the Concrete distribution which can be, as in the discrete case, free parameters $\mbphi = \{\bernrelaxedposterior_{ij}\}$ or have a smooth parameterization analogous to that in \cref{eq:smooth-param}, i.e.~$\bernrelaxedposterior_{ij} = \exp(\mbz_i^T  \mbz_j + b_i + b_j  + s)$ 
 %\todo{We don't need to restrict $0 < \lambda <1 $. Should this be the exponential transform?}
 and, consequently, $\mbphi=\{ \{\mbz_i, b_i\}, s \}$. 
%We note that, for simplicity, we have overloaded the notation on the location parameters of the binary Concrete distribution, which are also used in the Bernoulli posterior. 
\section{Binary discrete distributions}\label{sec:binary-discrete}
The \gls{KL} divergence between two Bernoulli distributions $q(a \g \berndiscreteposterior)$ and $p(a \g \berndiscreteprior)$ can be computed as
\begin{equation}
\KL{q(a \g \berndiscreteposterior)}{p (a \g \berndiscreteprior)}
= \berndiscreteposterior [\log \berndiscreteposterior - \log \berndiscreteprior ]
+ (1 - \berndiscreteposterior) [\log (1- \berndiscreteposterior) - \log (1 - \berndiscreteprior) ].
\end{equation}
\section{Binary concrete distributions}
\label{app:concrete-distros}
In this section we give details of the re-parameterization used for the implementation of our algorithm when both the prior and the approximate posterior are relaxed via the binary Concrete distribution \cite{maddison2016concrete,jang2016categorical}.
\subsection{Summary of Bernoulli relaxation transformations}
\label{sec:summary_of_bernoulli_relaxation_transformations}
\begin{align}
A_{ij} 
& 
\sim \mathrm{BinConcrete}(\bernrelaxedposterior_{ij}, \temposterior) 
\quad 
%& 
\Leftrightarrow 
%& 
\quad
A_{ij} = \sigmoid(B_{ij}),
\quad 
%& 
B_{ij} 
%& 
\sim \mathrm{Logistic} \left (\frac{\log \bernrelaxedposterior_{ij}}{\temposterior}, \frac{1}{\temposterior} \right ) ;
\\
B_{ij} 
& 
\sim \mathrm{Logistic} \left (\frac{\log \bernrelaxedposterior_{ij}}{\temposterior}, \frac{1}{\temposterior} \right ) 
\quad 
%& 
\Leftrightarrow 
%& 
\quad
B_{ij} = \frac{\log \bernrelaxedposterior_{ij} + L}{\temposterior},
\quad 
%& 
L 
%& 
\sim \mathrm{Logistic} (0, 1) ;
\\
L 
& 
\sim \mathrm{Logistic} (0, 1)
\quad 
%& 
\Leftrightarrow 
%& 
\quad
L = \sigmoid^{-1}(U) := \log U - \log(1 - U),
\quad 
%& 
U 
%& 
\sim \mathrm{Uniform} (0, 1).
\end{align}

In summary, we have
\begin{equation}
A_{ij} \sim \mathrm{BinConcrete}(\bernrelaxedposterior_{ij}, \temposterior) 
\quad \Leftrightarrow \quad
A_{ij} = \sigmoid \left ( \frac{\log \bernrelaxedposterior_{ij} + \sigmoid^{-1}(U)}{\temposterior} \right )
\quad U \sim \mathrm{Uniform} (0, 1).
\end{equation}
\subsection{Re-parameterized \gls{ELBO}}
With the results above, it is easy to see that we can write the \gls{ELBO} as:
\begin{align}
\cL_{\textsc{elbo}}(\mbphi) & = 
\E_{q_{\mbphi,\temposterior}(\mbA)} \left [ \log p_{\mbtheta}(\Yobs \g \mbX, \mbA) 
- \log \frac{q_{\mbphi,\temposterior}(\mbA)}{p_{\temprior}(\mbA)} \right ] \\
& = \E_{g_{\mbphi,\temposterior}(\mbB)} \left [ 
\log p_{\mbtheta}(\Yobs \g \mbX, \sigmoid(\mbB)) 
- \log \frac{q_{\mbphi,\temposterior}(\sigmoid(\mbB))}{p_{\temprior}(\sigmoid(\mbB))} \right ] \\
& = \E_{g_{\mbphi,\temposterior}(\mbB)} \left [ 
\log p_{\mbtheta}(\Yobs \g \mbX, \sigmoid(\mbB)) 
- \log \frac{g_{\mbphi,\temposterior}(\mbB)}{f_{\temprior}(\mbB)} \right ].
\end{align}
where
\begin{equation}
g_{\mbphi, \temposterior} (B_{ij}) = \Logistic\left(B_{ij} \g \frac{\log \bernrelaxedposterior_{ij}}{\temposterior}, \frac{1}{\temposterior}\right), \quad
 f_{\temprior} (B_{ij}) = \Logistic\left(B_{ij} \g \frac{\log \bernrelaxedprior_{ij}}{\temprior}, \frac{1}{\temprior}\right) .
\end{equation}

\subsection{Importance-weighted \gls{ELBO}}
\label{sec:iw-elbo}
For the relaxed version of our algorithm (that uses binary Concrete distributions), in which we cannot compute the KL term in the \gls{ELBO} analytically, we use the importance-weighted \gls{ELBO}, which has been shown to perform better than the standard \gls{ELBO}, be a tighter bound of the marginal likelihood and related to variational inference in an augmented space \citep{burda2015importance,domke-nips-2018}:
\begin{equation}
	\IWELBO =  \sum_{\mby_n \in \Yobs}  \LME_{\mbA_{1:S}} \left[     \log p( \mby_n | \mbX, \mbA ) - \frac{1}{| \Yobs|}  \log \frac{q_{\mbphi, \tau} (\mbA) } {p_{\tau_o}(\mbA)} \right] ,
\end{equation}
where $\LME_{\mbA_{1:S}}(h (\mbA))$ is the log-mean-exp operator of function $h(\mbA)$ over samples of $\mbA$, \ie $\LME_{\mbA_{1:S}} = \log \frac{1}{S} \sum_{s=1}^S \exp (h(\mbA^{(s)}))$ with $\mbA_{1:S} \equiv (\mbA^{(1)}, \ldots, \mbA^{(S)} )$ and $\mbA^{(s)} \sim q_{\mbphi, \tau} (\mbA)$.
\section{Implementation and computational complexity}
\label{sec:complexity-details}
%\todo[inline]{Need to check this section carefully that I haven't been too generous in lacking detail}
We implement our approach using TensorFlow \cite{tensorflow2015-whitepaper} for efficient GPU-based computation and also use some components of TensorFlow Probability \cite{dillon2017tensorflow}\footnote{Code available at \url{https://github.com/ebonilla/VGCN}.}. The time complexity of our algorithm can be derived from considering  the two main components of the \gls{ELBO} in \cref{eq:elbo-general}, namely the \gls{KL} divergence term and the \gls{ELL} term. We focus here on one-hidden layer \gls{GCN} (apart from the output layer)  with dimensionality $Q \equiv Q^{(1)}$  
along with a  smooth (dot product) parameterization of the posterior.
We recall  that $N$ is the number of nodes, $D$ is the dimensionality of the input features $\mbX$, $d_z$ is the dimensionality of posterior parameters $\mbZ$,  $C$ is the number of classes and $S$ is the number of samples from the variational posterior used to estimate the required expectations. 

% Additionally, we consider the relaxed and smooth parameterization of the posterior, \ie using the binary Concrete distribution along with its smooth (dot product) parameterization. This is the most computationally costly version of our algorithm but it yielded our best empirical results. 
%First of all, for estimating both the \gls{KL} and the \gls{ELL} we need to sample from our approximate posterior. Computing the parameterization of this posterior via the product $\mbZ \mbZ^T$ requires $\bigO(N^2 d)$ operations.  
% The \gls{KL} term requires summations over $\bigO(N^2)$ terms. 
\textbf{\gls{KL} divergence term}: 
We require to compute $\bigO(N^2)$  individual \gls{KL} divergences,  which can be trivially parallelized. In the case of the smooth parameterization, for both the discrete and the relaxed cases, we need to compute the dot-product between the latent representations for each each $A_{ij}$ which is $\bigO(d_z)$ and gradient information must be aggregated for each $\mbz_i, \tilde{\mbz}_i$.  While for a discrete posterior these individual \gls{KL} terms can be computed exactly (as shown in \cref{sec:binary-discrete}),  for the continuous relaxation we need to resort to  \gls{MC} estimation over $S$ samples.  Aggregation over samples can also be parallelized straightforwardly.

\textbf{\gls{ELL} term}: Computing the \gls{ELL} using a 2-layer \gls{GCN} as in \cref{eq:gcn-two-layers} requires $\bigO(NDQ + S(NQC + N^2Q + N^2C))$ for the continuous case. However, in the discrete case it only requires doing a forward pass over the standard \gls{GCN}  architecture $S$ times, hence being linear in the number of edges, \ie $\bigO(S |\mathcal{E}| D Q C)$, where $|\mathcal{E}|$ is the expected number of edges sampled from the posterior, assuming sparse-dense matrix multiplication is exploited.

%In terms of memory, our algorithm requires drawing from the approximate posterior over $\mbA$ explicitly. When using the binary Concrete distribution the leading factor is  $\bigO(N^2)$ as we do not require to keep the samples simultaneously to compute the empirical expectations. 

% \input{hierarchical.tex}
\section{Datasets}
\Cref{tab:datasets} gives details of the datasets used in our experiments. Training/Valid/Test refer to the default training/validation/test set sizes. However, as mentioned in \cref{sec:experiments}, we adopt a similar approach to that of \citet{lds-2019} where the training set  is augmented with approximately 50\% of the validation set.
\label{sec:datasets}
\begin{table}[t]
	\centering
	\caption{Datasets used in the experiments. Train/Valid/Test correspond the the training/validation/test set sizes. Label rate refers to the ratio of the training set size over the total number of nodes.}
	\scriptsize
	\begin{tabular}{llllllll}
		\toprule
		Dataset   & Type              & Nodes   & Edges     & Classes & Features & Train/Valid/Test & Label rate \\
		\midrule
		\cora     & Citation network  & 2,708   & 5,278     & 7       & 1,433 & 140/500/1,000  & 0.052      \\
		\citeseer & Citation network  & 3,327   & 4,676     & 6       & 3,703 & 120/500/1,000  & 0.036      \\
		\polblogs & Blog network      & 1,222   & 16,717    & 2       & N/A   & 122/275/825    & 0.10       \\
		\pubmed   & Citation network  & 19,717 & 44,338    & 3       & 500   &  60/500/1,000  & 0.003       \\
		\bottomrule
	\end{tabular}
	\label{tab:datasets}
\end{table}

% \begin{table}
%     \begin{minipage}{0.5\textwidth}
%         \caption{Datasets used in the experiments. Label rate refers to the proportion of labeled nodes used for training.}
%         \label{tab:datasets}
%         \scriptsize
%      	\begin{tabular}{lllllll}
%      		\toprule
%      		Dataset    & Nodes & Edges & Classes & Features & Label rate \\
%      		\midrule
%      		\cora      & 2,708 & 5,278   & 7       & 1,433    & 0.052      \\
%      		\citeseer  & 3,327 & 4,676   & 6       & 3,703    & 0.036      \\
%      		\polblogs  & 1,222 & 16,717  & 2       & N/A      & 0.10       \\
%      		\bottomrule
%      	\end{tabular}
%     \end{minipage}
%      \hfill
%      	\begin{minipage}{0.45\linewidth}
% 		\centering
% 	\end{minipage}
% \end{table}

\section{Full details of experimental set-up}
\label{sec:expt-setup}
Unless stated explicitly below, all the optimization-based methods were trained  up to a maximum of 5,000 epochs using the \adam optimizer  \cite{kingma2014adam}  with an initial learning rate of $0.001$. Hyper-parameter exploration was done via grid search and model selection carried out via cross-validation using the accuracy on the validation set. For standard \gls{GCN} and our method we use a two-layer \gls{GCN} as given in \cref{eq:gcn-two-layers} with a $16$-unit hidden layer.  We train standard \gls{GCN}  as done by \citet{kipf2016semi} so as to minimize the cross-entropy loss, using dropout and L2 regularization, Glorot weight initialization \cite{glorot2010understanding} and row-normalization of input-feature vectors.  As with our method, we set the dropout rate to $0.5$ and  the regularization parameter in $\{ 5 \times 10^{-3}, 5 \times 10^{-4} \}$. 
%
% Our method
%flags = {'--prior-type': 'knn',
%	'--posterior-type': posterior_type,
%	'--num-epochs': '5000',
%	'--initial-learning-rate': '0.001',
%	'--mc-samples-train': str(mc_samples_train),
%	'--mc-samples-test': str(mc_samples_test),
%	'--dropout-rate': '0.5',
%	'--layer-type': 'dense',
%	' --latent-dim': str(latent_dim),
%	'--logit-shift': str(logit_shift),
%	'--log-every-n-iter': '50',
%	'--zero-smoothing-factor': '1e-5',
%}
%settings = {'dataset': ['cora', 'citeseer'],
%	'method': [method],
%	'corruption': ['nograph'],
%	# 'prior': ['prior-free_lowdim'],
%	# 'init-size': ['init-size-10', 'init-size-100'],
%	# 'init-val': ['init-val-1e-3', 'init-val-1e-5'],
%	'prior': ['prior-knn'],
%	'knn-metric': ['knn-metric-cosine', 'knn-metric-minkowski'],
%	'knn-k': ['knn-k-10', 'knn-k-20'],
%	'regscale': ['regscale-5e-3', 'regscale-5e-4'],
%	'one-smoothing-factor': ['one-sf-0.25', 'one-sf-0.5', 'one-sf-0.75', 'one-sf-0.99'],
%	'temp-prior': ['temp-prior-0.1', 'temp-prior-0.5'],
%	'temp-posterior': ['temp-posterior-0.1', 'temp-posterior-0.5', 'temp-posterior-0.66'],
%	'beta': ['beta-1'],
%	'runs': SEED_VAL,
%}

For our method, we carried posterior estimation over the adjacency matrix and MAP estimation of the \gls{GCN}-likelihood parameters so as to maximize the \gls{ELBO} in \cref{eq:elbo-general}. Hyper-parameters for \gls{GCN}-estimation were the same as above. To construct the prior over the adjacency matrix we followed the procedure explained in \cref{sub:prior} with  $\smoothfactorone = \{0.25, 0.5, 0.75, 0.99\}$,  $\smoothfactorzero=10^{-5}$, $\temprior=\{0.1, 0.5\}$,  $\temposterior=\{0.1, 0.5, 0.66\}$ and $\beta=\{10^{-4}, 10^{-3}, 10^{-2}, 1 \}$. We initialized the posterior to the same smoothed probabilities in the prior and used   $S_{\text{train}}=3$ and $S_{\text{pred}}=16$ samples  for estimating the required expectations for training and predictions, respectively.  In the no-graph case all the methods explored \glspl{KNNG} with $k=\{10, 20\}$ and distance metrics $\{\text{cosine}, \text{Minkowski} \}$.

% LDS
For \gls{LDS} we used  the code provided by the authors\footnote{\url{https://github.com/lucfra/LDS-GNN}.},  which carries out bilevel optimization of the regularized cross-entropy loss and does model selection based on the validation accuracy using grid search across a range of parameters such as learning rates (for inner and outer objectives), number of neighbors and distance metrics. Similarly, for \gls{GGP} and \gls{EGCN} we used the code provided by the authors\footnote{\url{https://github.com/yincheng/GGP} for \gls{GGP} and \url{https://github.com/huawei-noah/BGCN} for \gls{EGCN}.}. 

For \gls{RGCN} we used the code provided by the authors\footnote{\url{https://zw-zhang.github.io/files/2019_KDD_RGCN.zip}}. For all experiments, we used the default parameters as described in \citep{robustgcn2019}. Specifically, we used 2 hidden graph convolutional layers with $32$ units each and dropout 0.6. All models were trained for a maximum $200$ epochs with early stopping (20 epochs patience) using the \adam optimizer and an initial learning rate of $0.01$. 

We used our own implementation of \gls{GAT} and \gls{GRAPHSAGE}. \gls{GAT} used an architecture identical to the one described in \citet{velickovic2018graph}. The first layer consists of $K= 8$ attention heads computing $F= 8$ features, followed by an exponential linear unit (ELU) nonlinearity. The second layer is used for classification that computes $C$ features (where $C$ is the number of classes), followed by a softmax activation. $L2$ regularization with $\lambda= 0.0005$ and $0.6$ dropout was used. The implementation of \gls{GRAPHSAGE} used mean aggregator functions and sampled the neighborhood at a depth of $K=2$ with neighborhood sample size of $S_1=25, S_2=10$ and batch size of  $50$. The model was trained with 0.5 dropout and $L2$ regularization with $\lambda= 0.0005$.

\section{Complete set of results for the adversarial setting}
\label{sec:allattackedgraphs}
\begin{figure*}[t]
	\centering
	\includegraphics[width=0.95\textwidth]{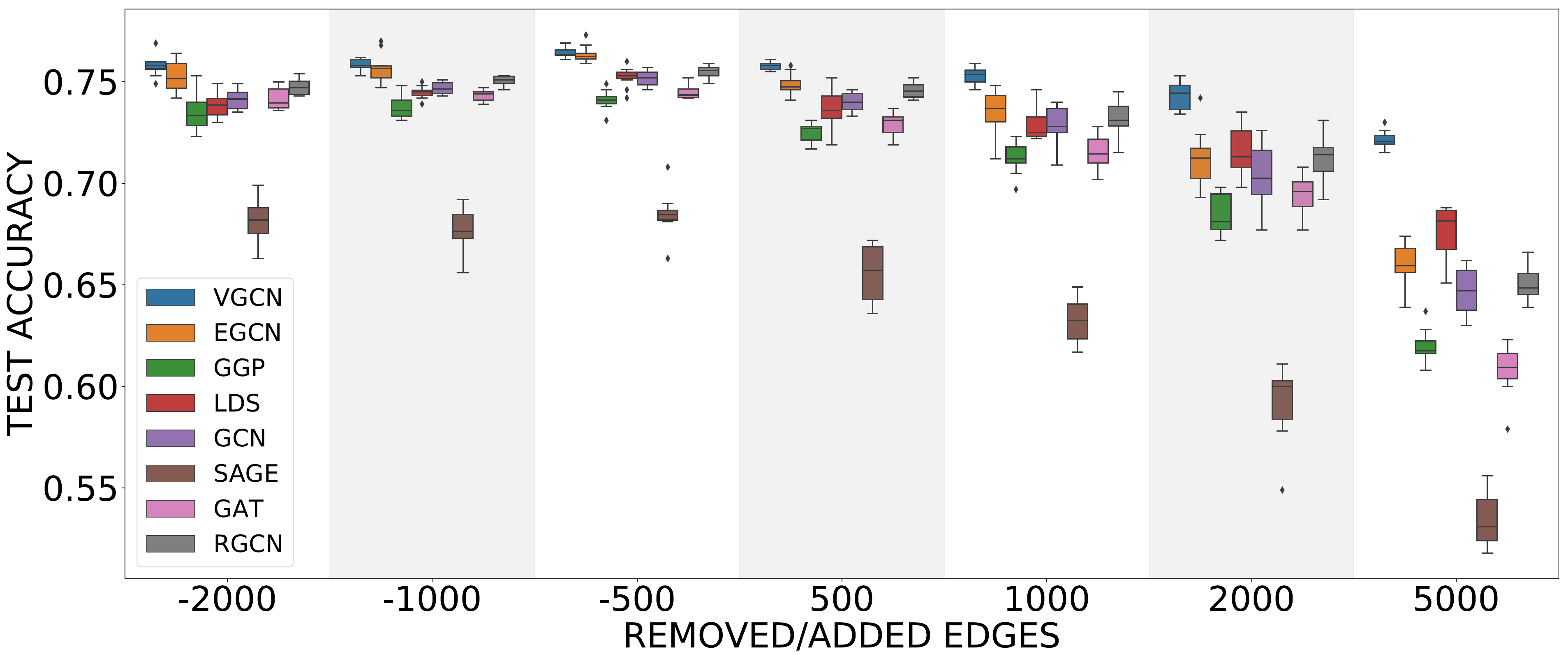}
	\includegraphics[width=0.95\textwidth]{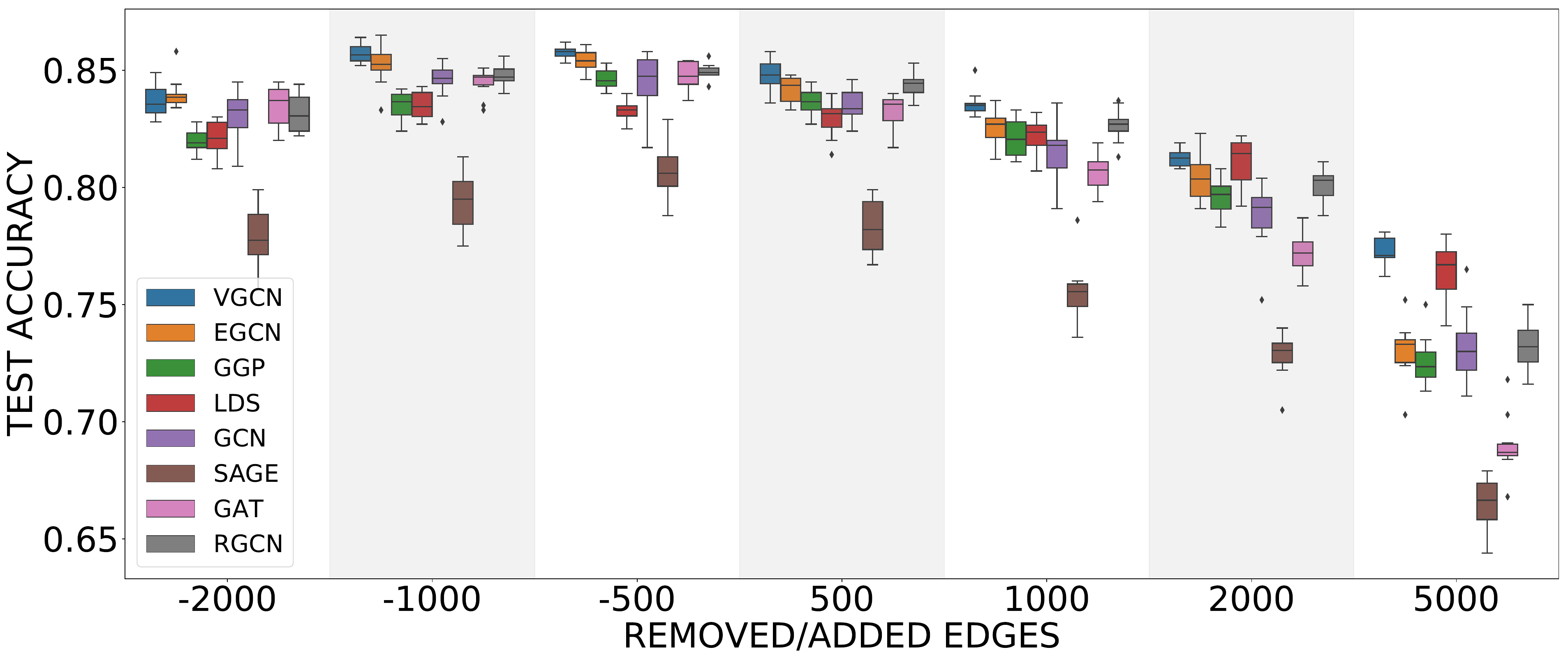}
	\includegraphics[width=0.95\textwidth]{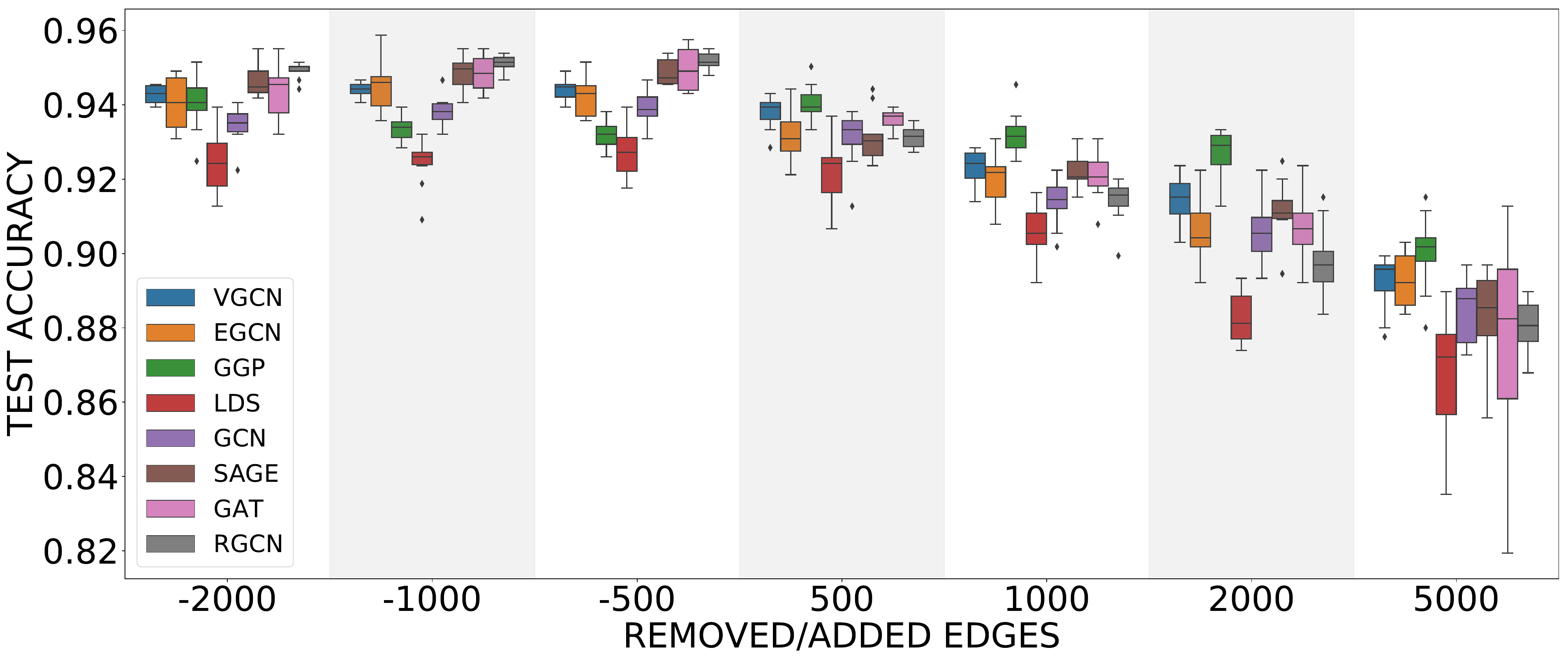}
	\caption{Results for the adversarial setting on attributed graphs \citeseer (top),  \cora (middle), and (featureless graph) \polblogs (bottom): Accuracy on the test set across ten attacked graphs at each attack setting such that negative values indicate removing edges and positive values adding edges. We compare our method (\gls{VGCN}) with competing algorithms.
	\label{fig:allattacked}}
\end{figure*}

Here we include the complete set of experiments for the 7 attacked graphs removing 2000, 1000, and 500 edges as well as adding 500, 1000, 2000, and 5000 edges to the ground truth graphs. Figure~\ref{fig:allattacked} shows the results for all graphs. 

On the citation networks, \citeseer and \cora, our proposed  \gls{VGCN} outperforms all other Bayesian and non-Bayesian methods, especially in the case of adding edges. On the \polblogs network that is lacking node features, all methods perform similarly with \gls{GGP} having a small edge in the cases of adding 2000 and 5000 edges. Our method, displays the lowest variance across all datasets.

%\section{Additional results}
%Here we present additional results analyzing different settings for our model (\gls{VGCN}). 
%
\section{Low-rank vs free parameterizations}
\begin{figure}
	\centering
	\includegraphics[width=0.45\textwidth]{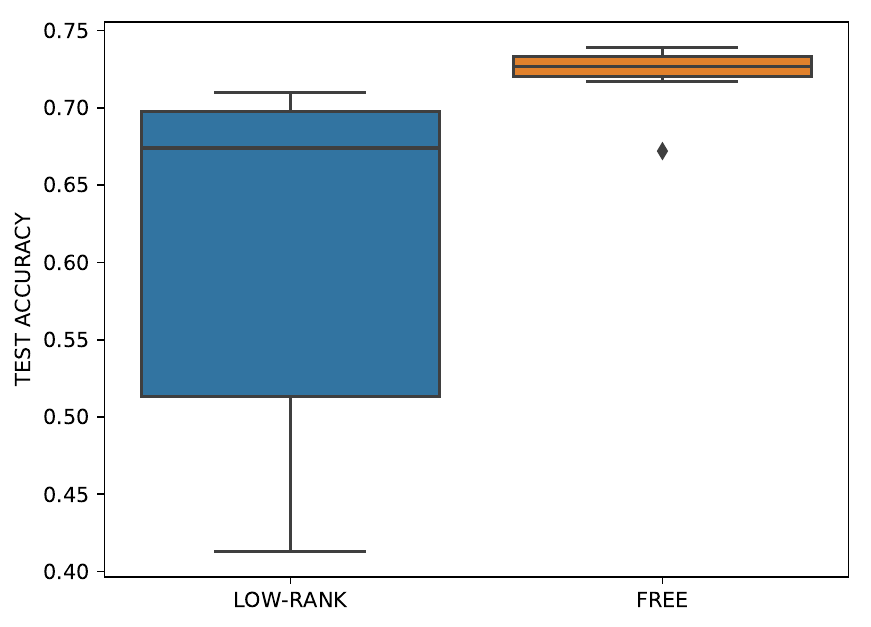}
	\includegraphics[width=0.45\textwidth]{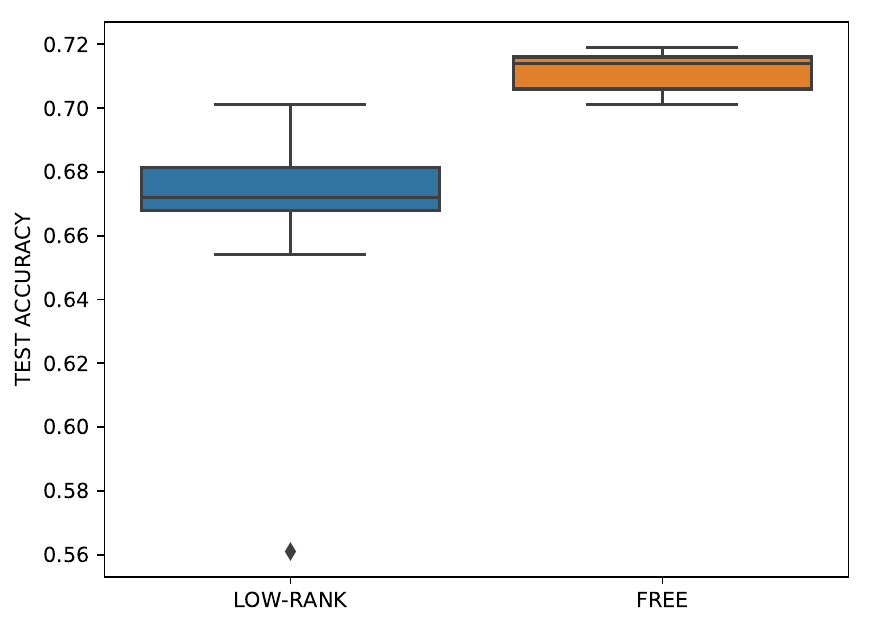}
	\caption{Results in the no-graph case for \citeseer (left) and \cora (right)---low-rank vs free parameterization: The test accuracy of our method (\gls{VGCN}) using a low-rank parameterization and a free parameterization, with the latent dimensionality of the low-rank parameterization $d_z = 100$. As with the results in the main paper, In both cases we used $S_{\text{pred}} = 16$ posterior samples for predictions. \label{fig:lowrank-vs-free}}
\end{figure}

\Cref{fig:lowrank-vs-free} compares the low-rank parameterization vs the free parameterization of our model, where we used a latent representation of dimensionality $d_z = 100$. Our goal is to analyze whether a much more compact representation can yield similar results to those obtained by the free parameterization. For the citation networks used, the number of latent  variables with the low-rank parameterization is $N \times 2 \times d_z \approx 6 \times 10^5$, whereas with the free parameterization we have $N \times (N-1)/2 \approx 4.5 \times 10^6$ latent variables. i.e.~in this setting, the low-rank parameterization has an order of magnitude fewer latent variables. We see, in \cref{fig:lowrank-vs-free}, that although the low-rank parameterization can in some cases achieve a performance close to that of the free-parameterization, it also has a much higher variance and in most cases the resulting solution is considerably poorer. Nevertheless, we believe that factors such as initialization can improve the performance of the low-rank parameterization significantly and leave a much more thorough study of this for future work.

\section{Discrete vs relaxed}
\begin{figure}
	\centering
	\includegraphics[width=0.45\textwidth]{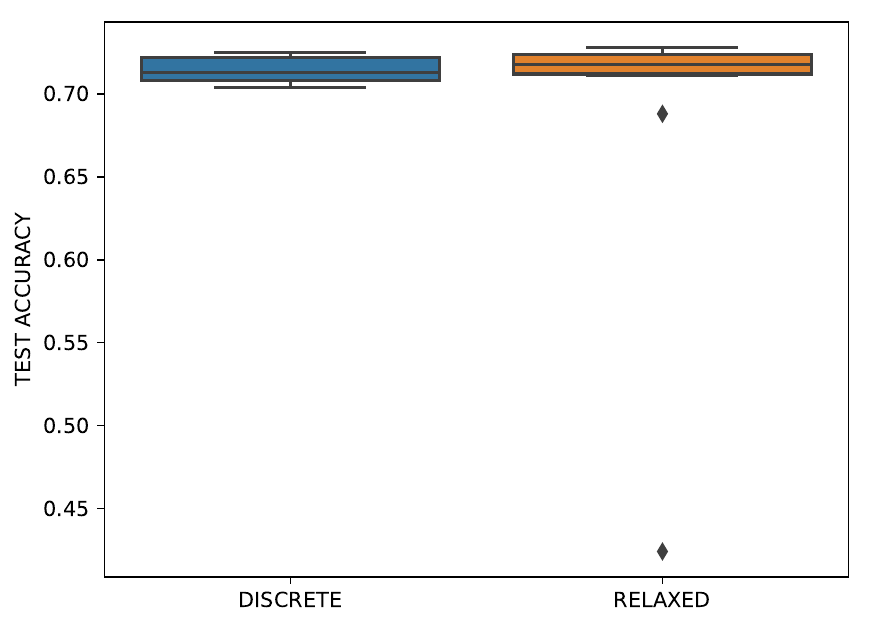}
	\includegraphics[width=0.45\textwidth]{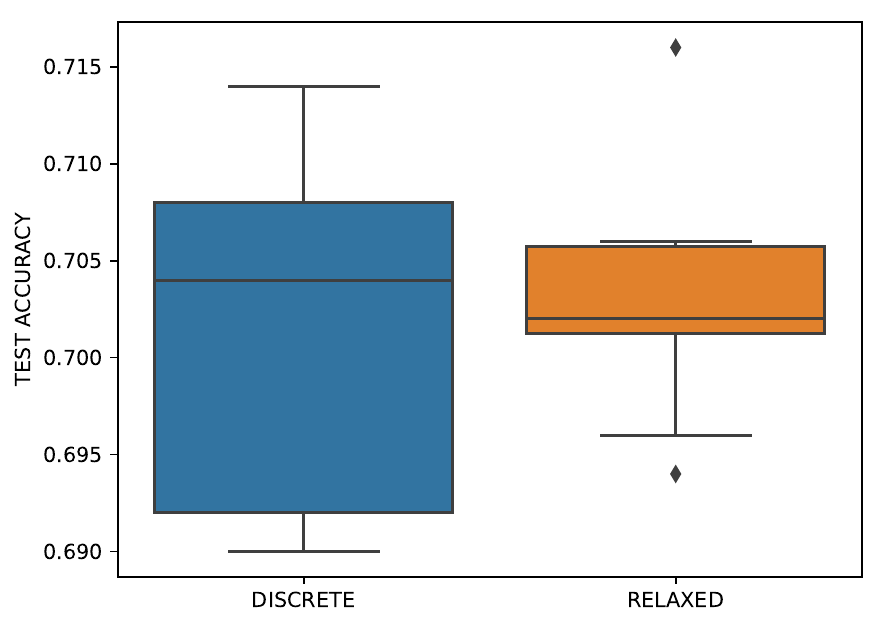}
	\caption{Results in the no-graph case for \citeseer (left) and \cora (right)---discrete vs relaxed: The test accuracy of our method (\gls{VGCN}) using a free parameterization with discrete Bernoulli distributions and relaxed binary Concrete distributions. Unlike the results in the main paper, in both cases we used $S_{\text{pred}} = 3$ posterior  samples for predictions. \label{fig:discrete-vs-relaxed}}
\end{figure}

\begin{figure}
		\includegraphics[width=0.5\textwidth]{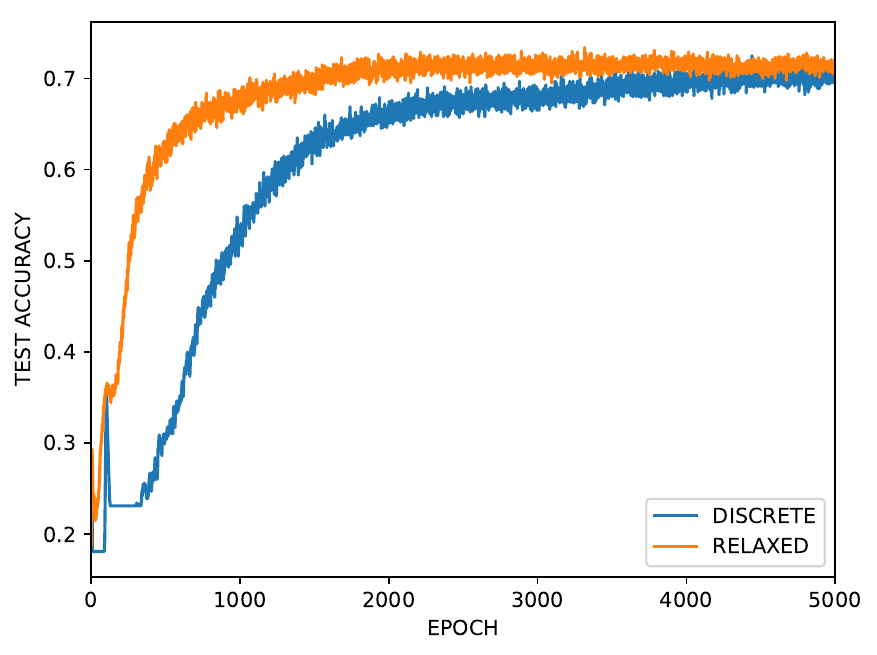}
		\caption{Results in the no-graph case---convergence of discrete vs relaxed approaches on \citeseer: Convergence of the free parameterization with the discrete and the relaxed version using $S_{\text{pred}} = 3$ posterior  samples for predictions. \label{fig:convergence}}
\end{figure}

Besides the number of latent variables used to represent the posterior, we also want to investigate the effect of using the discrete Bernoulli distributions along with the score function estimators versus the relaxed binary Concrete distributions and the reparameterization trick. \Cref{fig:discrete-vs-relaxed} shows the performance of these two approaches on the citations networks under study and the no-graph case when using only $S=3$ posterior samples for prediction. We see that there is not much difference between the two approaches, although the relaxed version exhibits some outliers on \citeseer (which is ameliorated when using $S=16$ samples) and the discrete version has slightly higher variance on \cora. However, as we see in \cref{fig:convergence}, the relaxed version converges much faster than the discrete version, hence our selection of the former for our main results.

\begin{figure}
	\centering
	\includegraphics[width=0.99\textwidth]{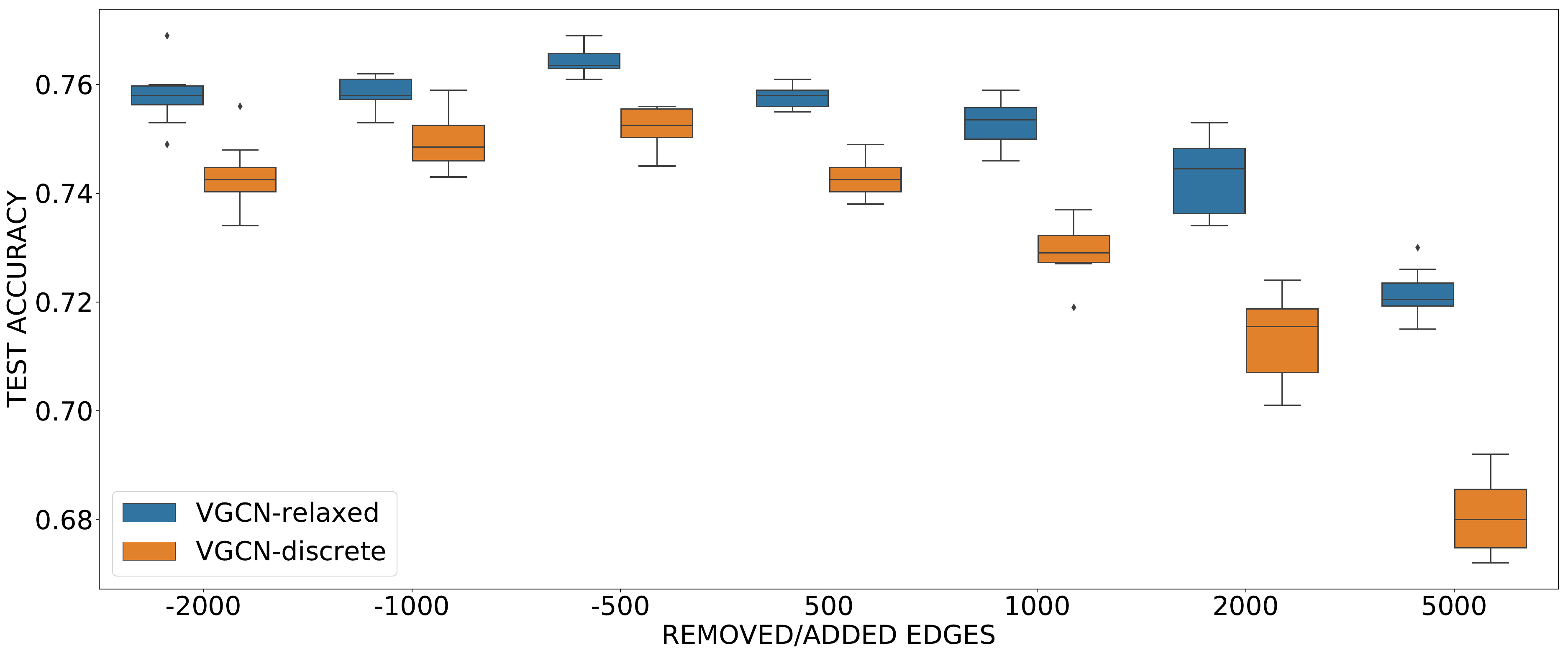}
	\includegraphics[width=0.99\textwidth]{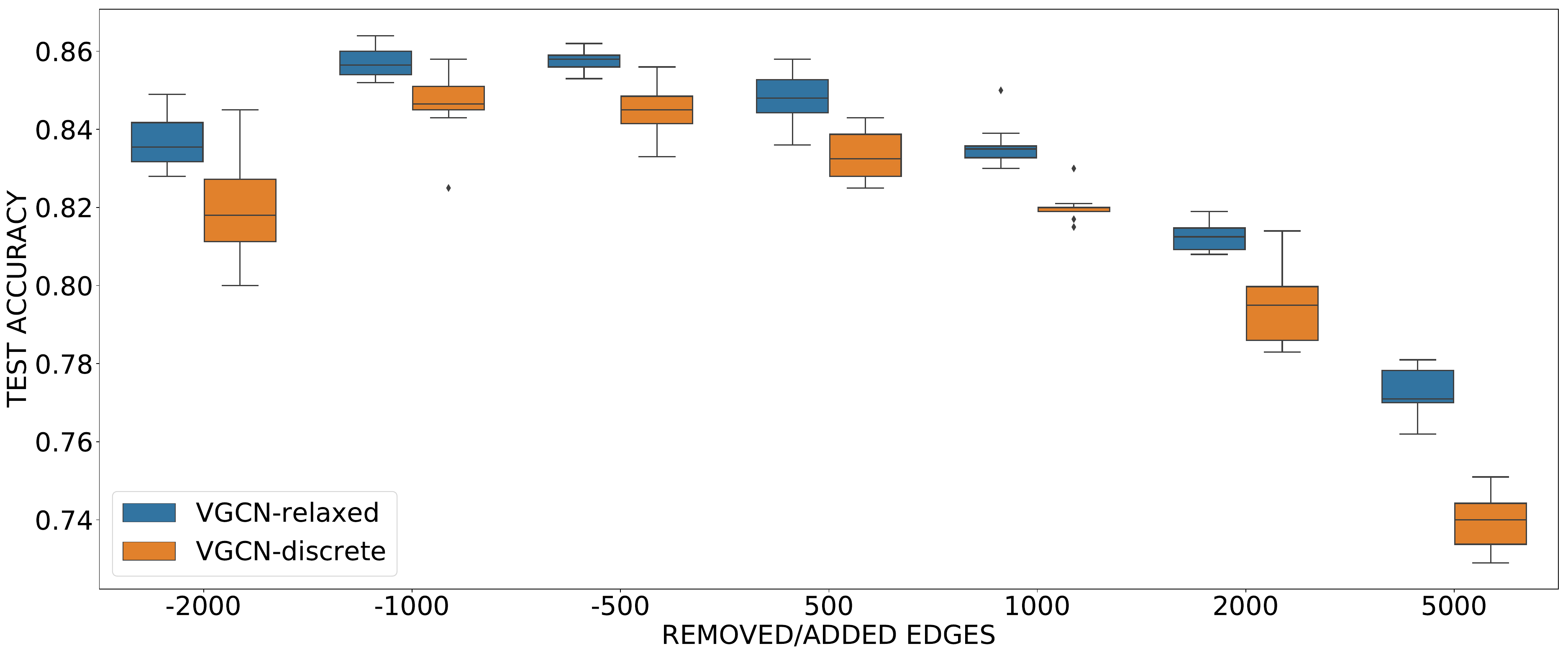}
	\caption{Results in the adversarial setting for \citeseer (top) and \cora (bottom) ---discrete vs relaxed: The test accuracy of our method (\gls{VGCN}) using a free parameterization with discrete Bernoulli distributions (\gls{VGCN}-discrete) and relaxed binary Concrete distributions (\gls{VGCN}-relaxed). In both cases we used $S_{\text{pred}} = 3$ posterior  samples for predictions. \label{fig:attacked-discrete-vs-relaxed}}
\end{figure}

\Cref{fig:attacked-discrete-vs-relaxed} shows results of the discrete vs relaxed  parameterization in the adversarial setting for the \citeseer and \cora datasets. Both models were trained for a maximum 5000 epochs with hyper-parameter optimization and model selection as described in \cref{sec:expt-setup}. In the adversarial setting, there is an advantage to using the relaxed parameterization as it clearly outperforms the discrete one across all attack settings and both datasets. The difference in performance is more pronounced for \citeseer. Lastly, the relaxed parameterization exhibits lower variance across all graphs.

\section{The Effect of the number of posterior samples on predictions}
\begin{figure} 
	\centering
	\includegraphics[width=0.45\textwidth]{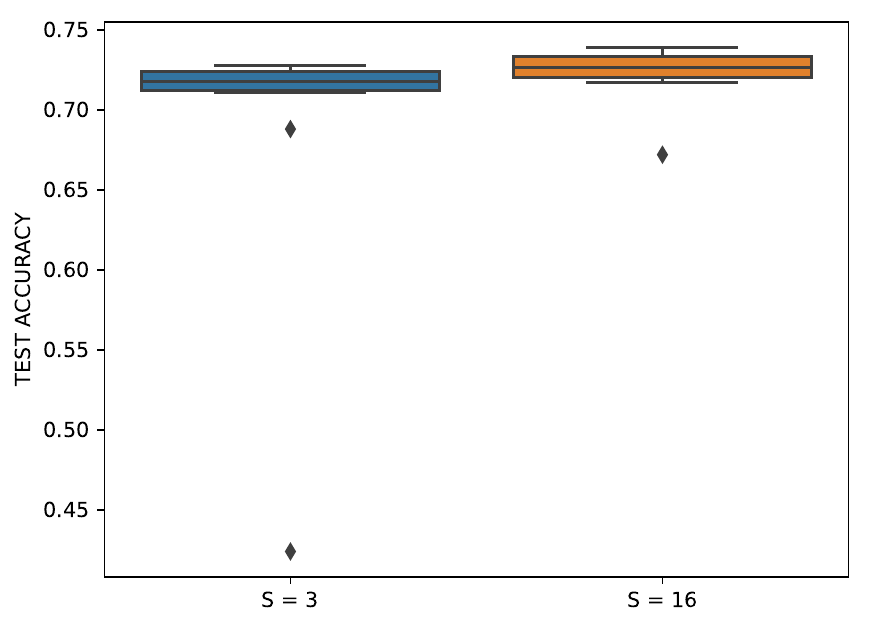}
	\includegraphics[width=0.45\textwidth]{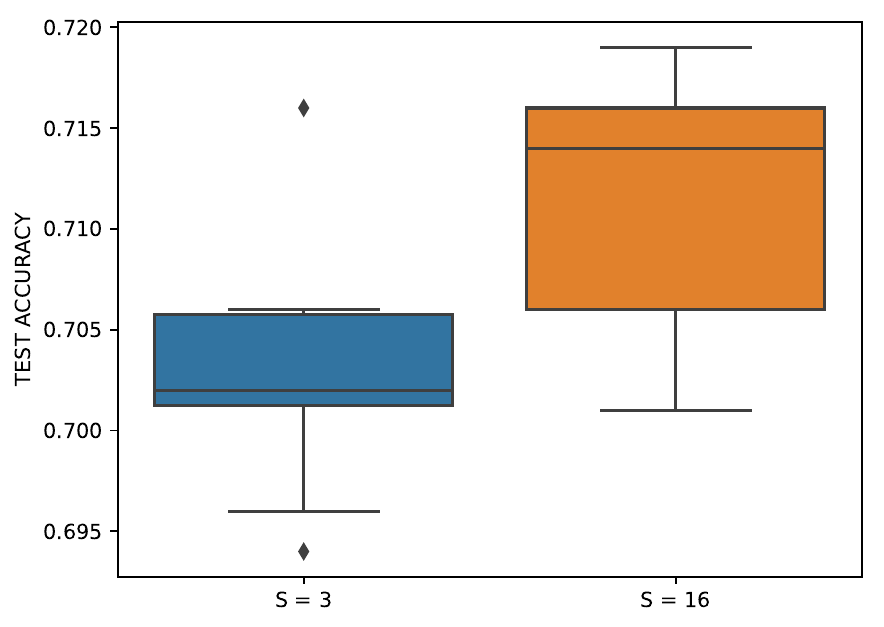}
	\caption{Results in the no-graph case for \citeseer (left) and \cora (right)---effect of the number of samples: The test accuracy of our method (\gls{VGCN}) (with freely parameterized relaxed binary Concrete distributions) using  $S \equiv S_{\text{pred}}=3$ and $S \equiv S_{\text{pred}} = 16$ posterior samples for predictions. \label{fig:samples}}
\end{figure}

\Cref{fig:samples} shows results for our model when using $S=3$ and $S=16$ samples from the posterior when making predictions. We observe that, as expected, using more posterior samples does improve performance and that the additional gains of using more samples are worthwhile if the computational constrains can be satisfied.

\section{The influence of the KL term during training}
\begin{figure}
	\centering
	\includegraphics[width=0.45\textwidth]{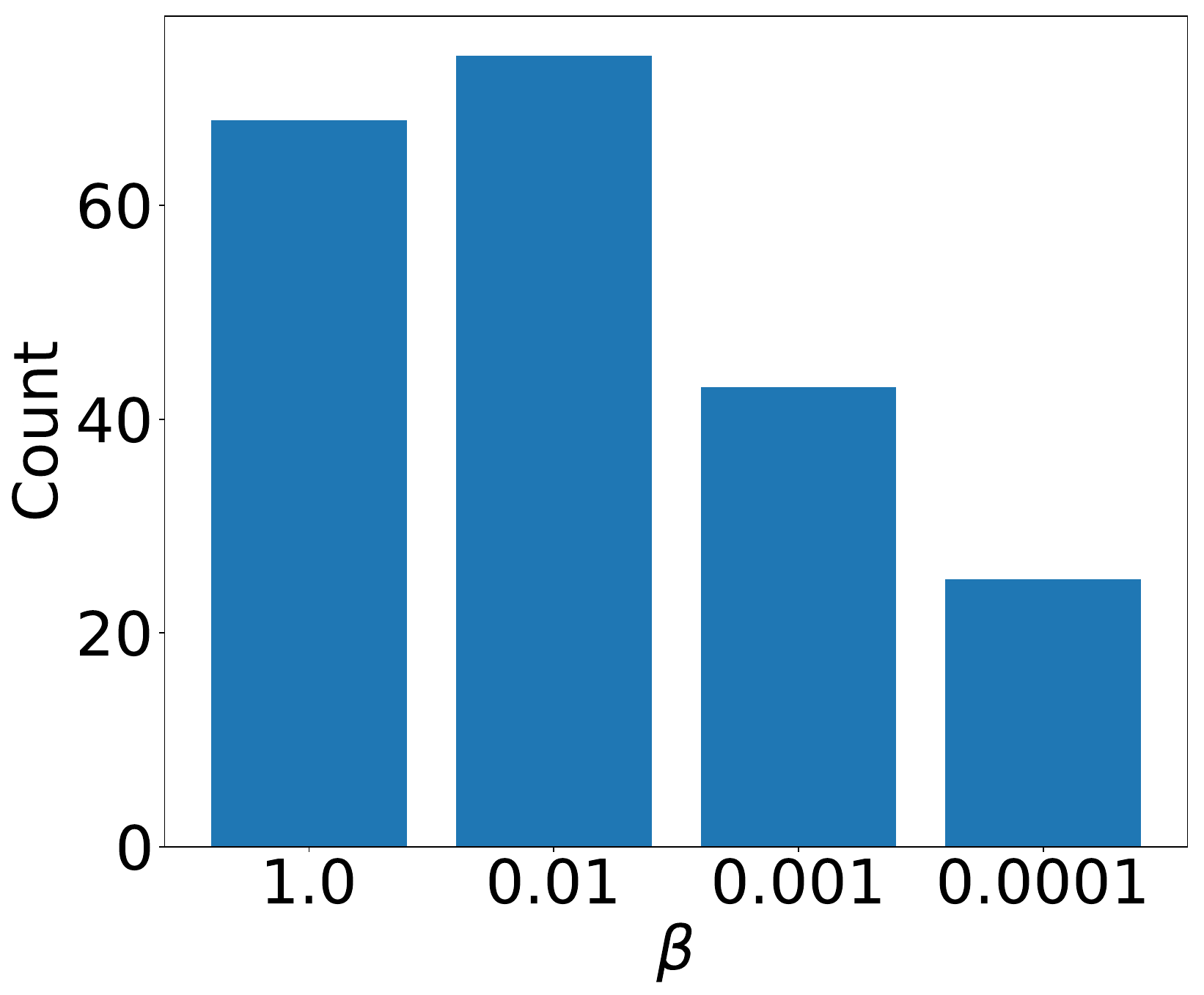}
	\caption{Histogram of \gls{KL}-dampening factor $\beta$  selected using cross-validation for experiments using the \citeseer, \cora, and \polblogs datasets, where $\beta=1$ accounts for no dampening of the \gls{KL} term in the \gls{ELBO}.
	\label{fig:beta-histogram}}
\end{figure}

As mentioned in the main paper, we scaled the \gls{KL} term by a dampening factor $\beta <1$ so that it does not dominate the likelihood term in the \gls{ELBO}. We have analyzed this KL-dampening factor on the adversarial experiments across all datasets.  \Cref{fig:beta-histogram} shows how frequently each factor was selected through cross-validation with  $\beta=\{1, 10^{-2}, 10^{-3}, 10^{-4}\}$ being selected $\{32\%,35\%, 21\%, 12\% \}$ of the time, respectively. Along with the performance benefits shown in the main paper, this confirms that KL regularization resulting from variational inference does have an effect. Lastly, we have found that even when $\beta$ is very small, the \gls{KL} term still has an effect as it can be several orders of magnitude larger than the \gls{ELL}.
\section{Posterior analysis in the adversarial setting}
\begin{figure}
		\centering
		\includegraphics[width=0.45\textwidth]{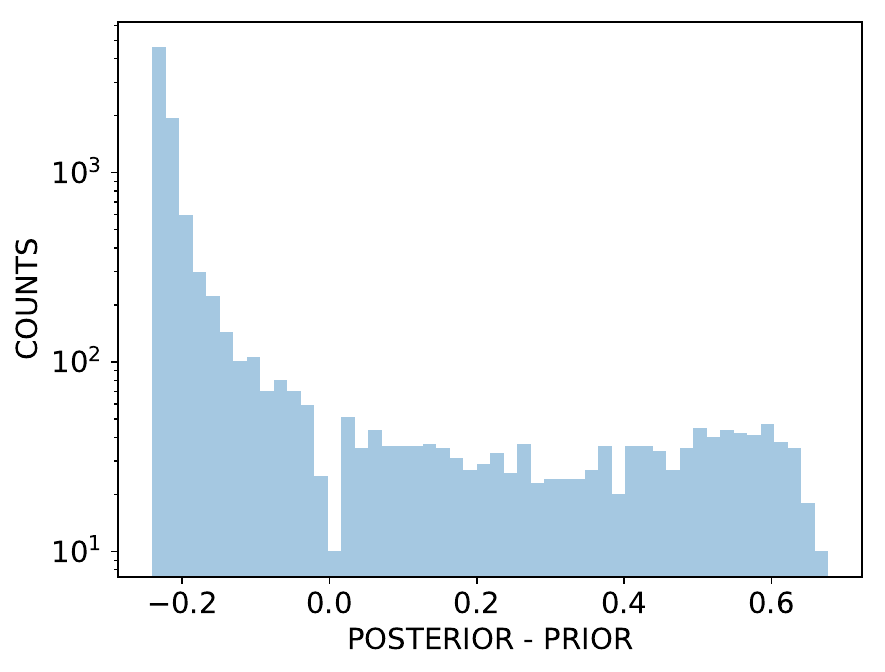}
		\caption{Posterior changes in the adversarial setting on \citeseer. The difference in the limit probabilities computed as the zero-temperature limit of the final variational posterior distributions over each adjacency entry. Only showing those probabilities that changed significantly from the prior, which had a maximum value of $0.25$ and a minimum value of $10^{-5}$. The total number of changed probabilities was $9,504$. \label{fig:posterior-adv} }
\end{figure}

%PYTHONPATH=. python experiments/gcn_bayesian.py --dataset-name=citeseer  --initial-learning-rate=0.001 --dropout-rate=0.5 --l2-reg-scale-gcn=0.0005 --layer-type=dense --posterior-type=free --one-smoothing-factor=0.25 --load-adj-matrix=attacked_datasets/citeseer/citeseer_attack_add_5000_v_6.gpickle --num-epochs=5000 --beta=0.01 --temperature-posterior=0.5 --temperature-prior=0.1 --use-half-val-to-train  --results-dir=./test_attack --save-checkpoint-steps=1000  --checkpoint-dir=results/test_attack/chkpt bayesGCN
Similar to the analysis in  \cref{sec:posterior-analysis} of the main paper for the no-graph case, here we look at the posterior changes for a representative experiment in the adversarial setting.  \Cref{fig:posterior-adv} shows the  difference between the final posterior probabilities obtained by our algorithm and the prior probabilities, which were also used to initialize the posterior.  We see that our model manages to effectively turn off/turn on a significant number of links. %Overall, the number of links that changed significantly, as defined in the main paper, was of $9, 504$. 
\section{Additional examples of learned graphs}
\label{sec:communities}

\begin{figure}[t]
  \centering
  \includegraphics[width=0.38\textwidth]{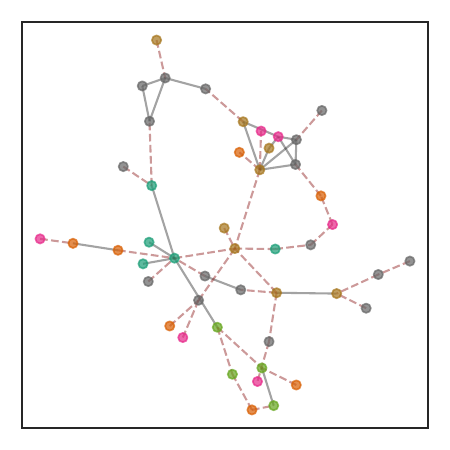}
  ~
  \includegraphics[width=0.38\textwidth]{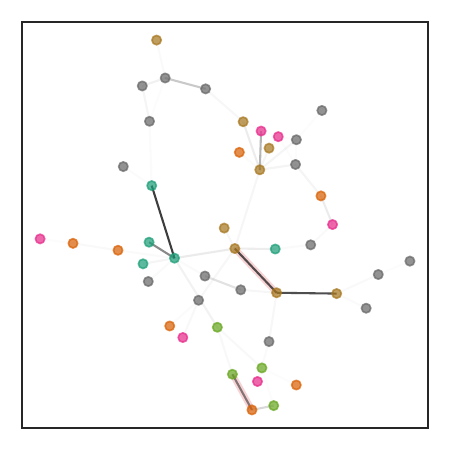}
  
  \includegraphics[width=0.38\textwidth]{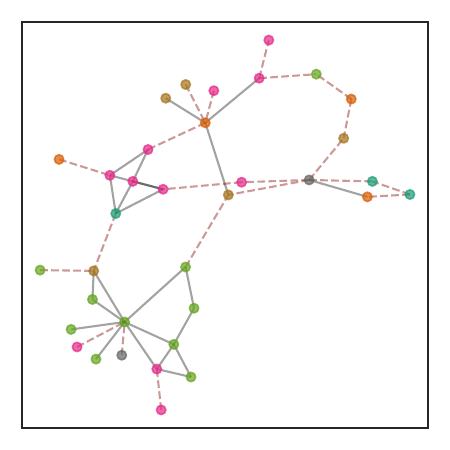}
  ~
  \includegraphics[width=0.38\textwidth]{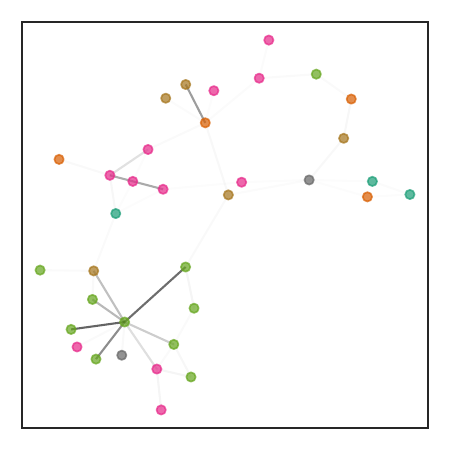}
  
  \includegraphics[width=0.38\textwidth]{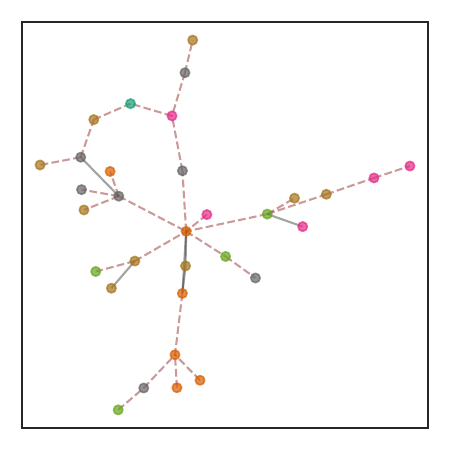}
  ~
  \includegraphics[width=0.38\textwidth]{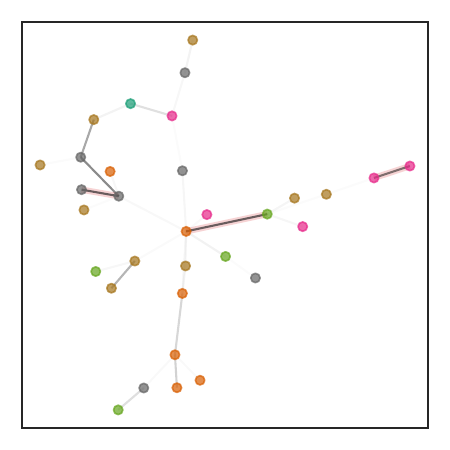}
  
  \includegraphics[width=0.38\textwidth]{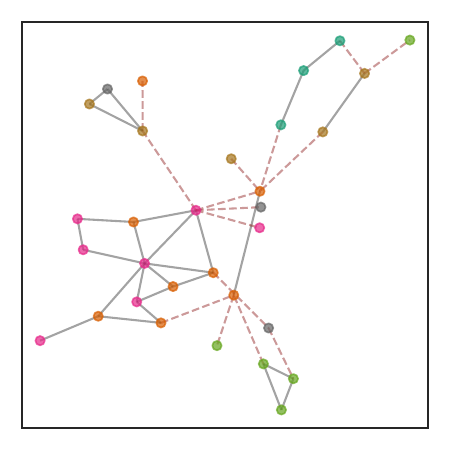}
  ~
  \includegraphics[width=0.38\textwidth]{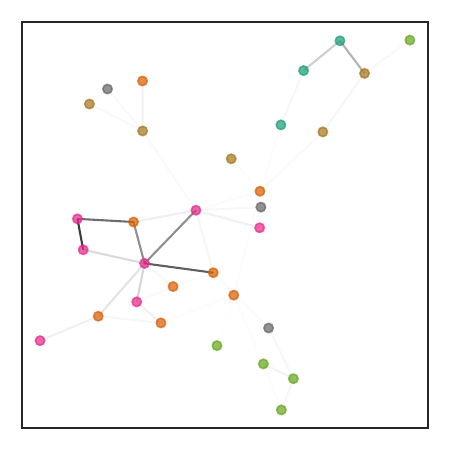}
  \caption{\emph{Left:} Communities in the original graph from the adversarial \citeseer experiment with node labels distinguished by colors and added edges denoted by red dashes. 
  \emph{Right:} Learned graph with edge opacity proportional to limit posterior probabilities. Added edges with probability greater than \nicefrac{1}{2} are highlighted red.} 
  \label{fig:communities}
\end{figure}

Following on from \cref{sec:qualitative_analysis} of the main paper, we 
visualize additional communities of the \textsc{citeseer} citation network
with added edges and the corresponding latent graph inferred using our approach.
We show four communities in \cref{fig:communities}. 
As before, on the left we denote the edges from the original graph in solid 
lines and the added edges in dashed red lines; on the right is the 
corresponding complete graph with edge opacity drawn proportionally to the 
limit posterior probabilities. 

Furthermore, let $E'$ denote the set of added edges. If we have inferred a 
posterior that suppresses the negative influence of these added edges, we would 
expect that $\{i, j\} \in E'$ implies $p(A_{ij}) < \nicefrac{1}{2}$. 
% material implication means this is equivalent to:
% $\{i, j\} \notin E' OR p(A_{ij}) < 0.5$. 
% so negating all of this leads to
% $\{i, j\} \in E' and p(A_{ij}) \geq 0.5$. 
On the right, we highlight in red every edge $\{i, j\} \in E'$ where
$p(A_{ij}) \geq \nicefrac{1}{2}$. We can see that such cases are few and far 
between, even in communities predominantly consisting of added edges (e.g. row 3).

\section{Results using random data splits}
\begin{figure}[t]
	\centering
	\includegraphics[width=0.95\textwidth]{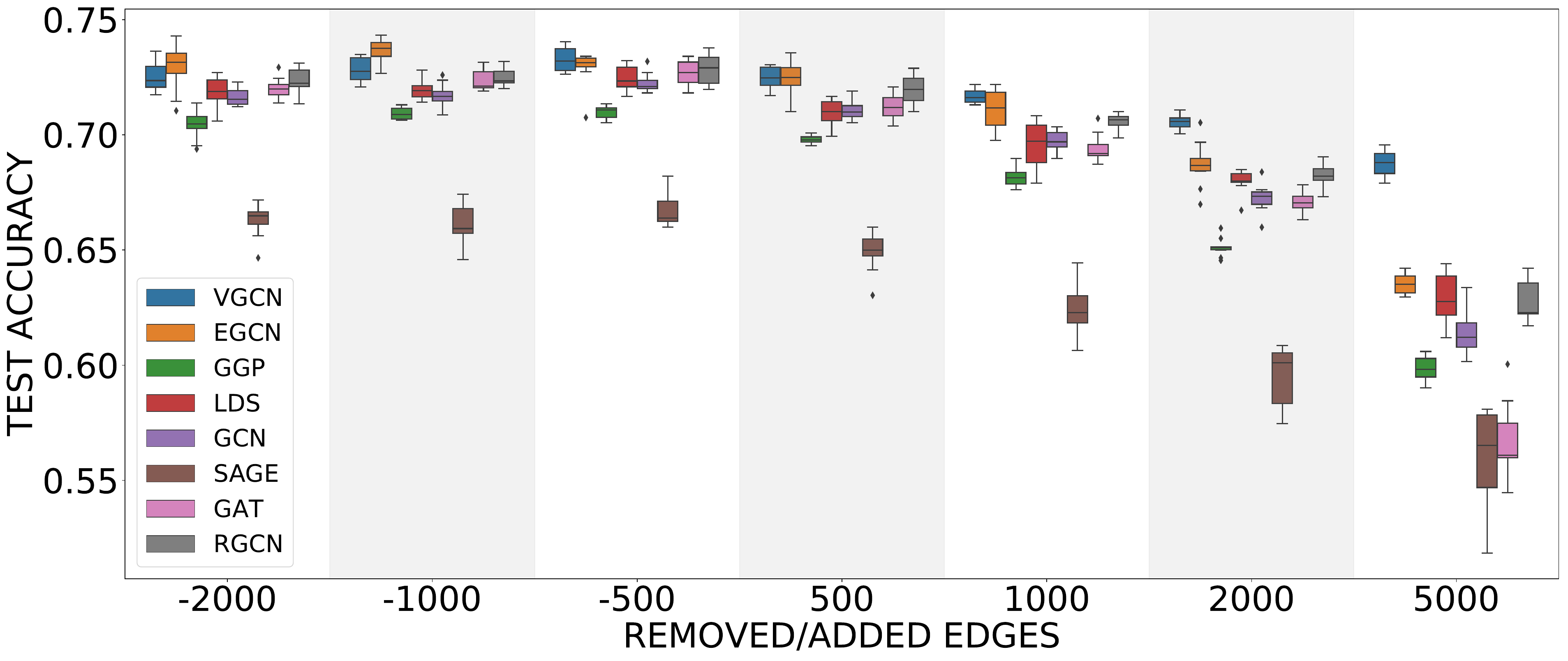}
	\includegraphics[width=0.95\textwidth]{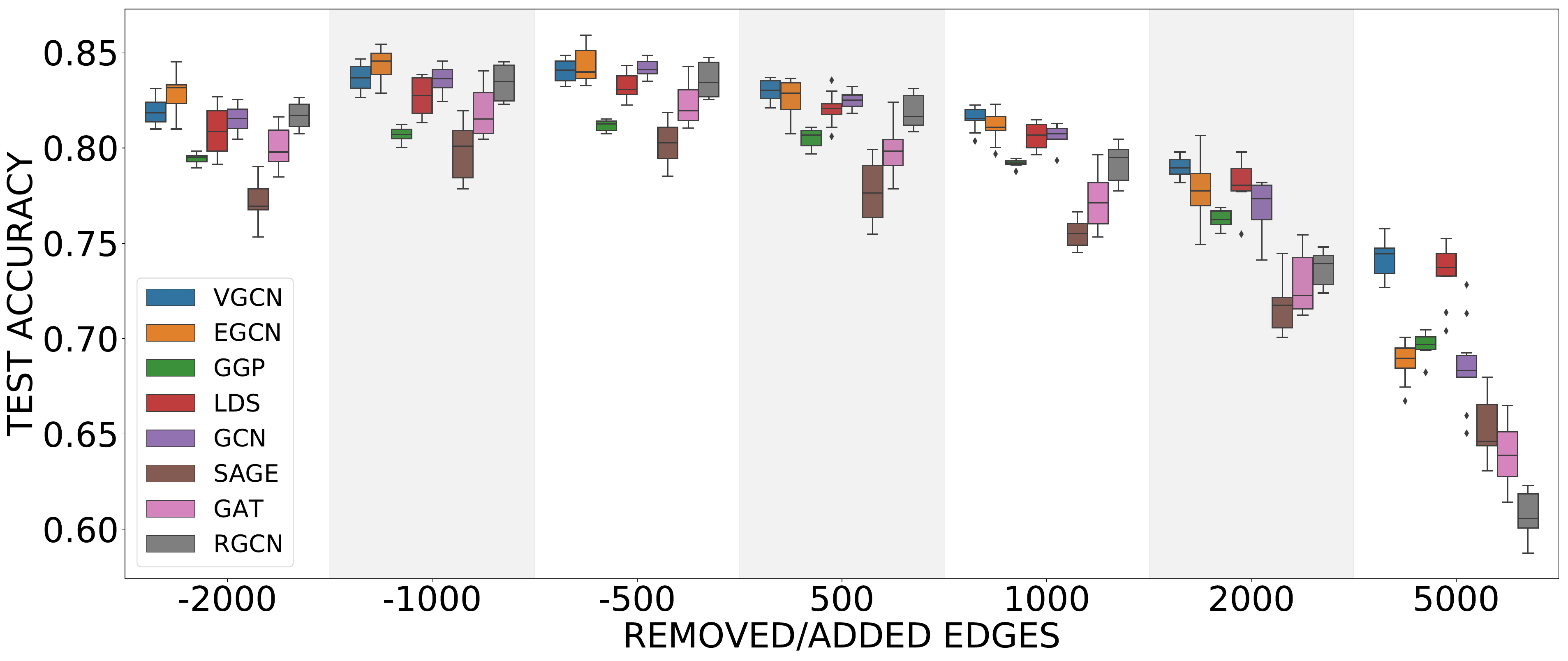}
	\caption{Results for the adversarial setting on attributed graphs \citeseer (top),  \cora (bottom) using random splits of the data into train/validation/test sets: Accuracy on the test set across ten attacked graphs at each attack setting such that negative values indicate removing edges and positive values adding edges. We compare our method (\gls{VGCN}) with competing algorithms.
	\label{fig:allattacked-random-split}}
\end{figure}

It was argued in \cite{shchur2018pitfalls} that model evaluation using pre-existing data train/validation/test splits produces overconfident estimates of a GNN model's performance. It is thus suggested that random splits of the data should be instead used. Here we have repeated the experiments outlined in section ~\ref{sec:allattackedgraphs} using random splits and show the results in \cref{fig:allattacked-random-split}. We have used the same random splits to evaluate all competing methods. 

Comparing the results shown in~\cref{fig:allattacked} and~\cref{fig:allattacked-random-split}, we notice that there is a small drop in performance for all methods and across all attack settings. However, overall our \gls{VGCN} method continues to outperform the others especially in the setting of adding a large number of false edges. The \gls{EGCN} method has a small advantage over \gls{VGCN} when removing $2000$ and $1000$ edges on both datasets; however, we note that the variance of \gls{EGCN} has also increased considerably when compared to the results using the fixed splits as shown in~\cref{fig:allattacked}. Overall, our original conclusions about the benefits of  \gls{VGCN}  in the case of attacked graphs remain true regardless of how the given data is split for training and validation.

\section{Mini-batch training}
\label{sec:mini-batch}
We explored mini-batch training employing the approach of \cite{ClusterGCN}. Mini-batch training permits applying our method to larger datasets where the full-batch method would result in out-of-memory errors. An additional benefit is also an order reduction in the number of model parameters through a block-diagonal approximation of the given graph; this approach is referred to as vanilla cluster-\gls{GCN} in \cite{ClusterGCN}. We decompose the auxiliary graph into $m$ non-overlapping subgraphs using the METIS \cite{metis} algorithm. Then, we optimise the \gls{VGCN} parameters using mini-batch SGD considering each subgraph as a mini-batch.

\begin{figure}
	\centering
	\includegraphics[width=0.7\textwidth]{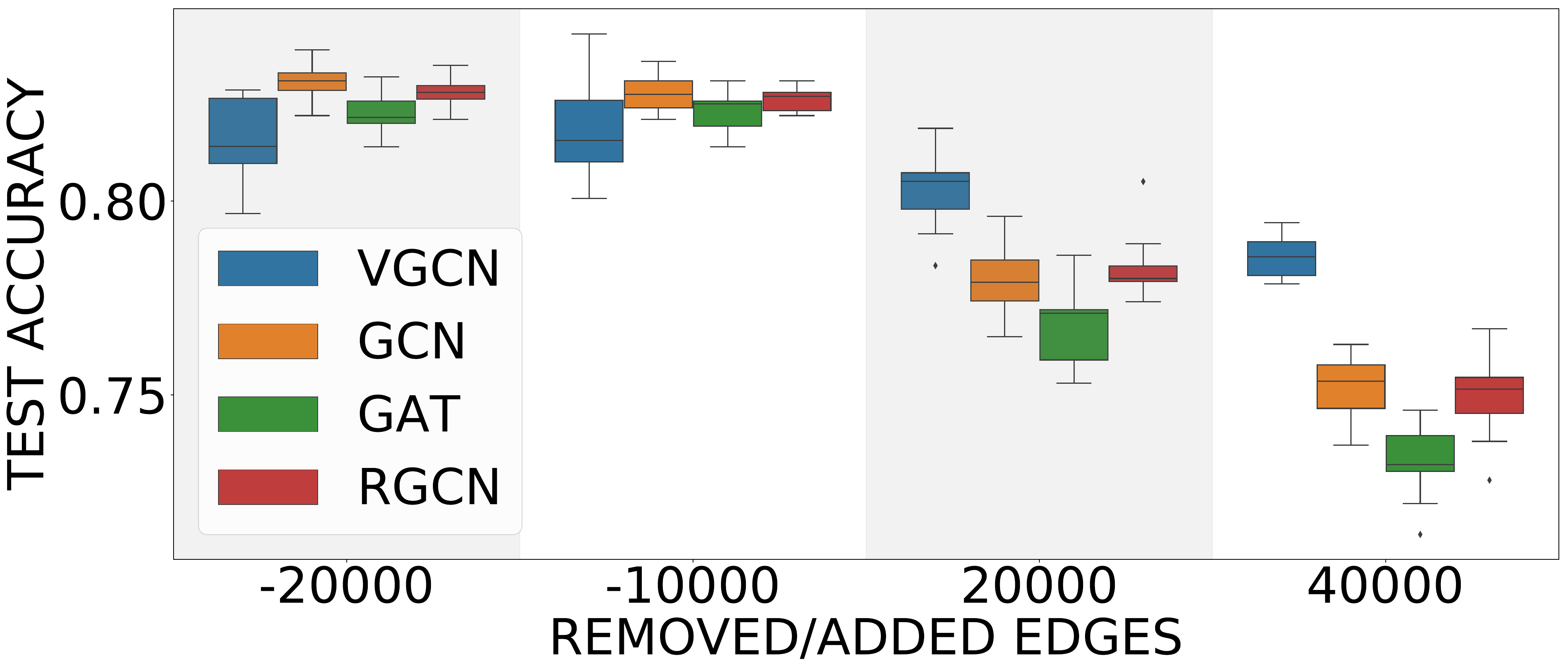}
	\caption{Results using mini-batch training with a block-diagonal approximation of the auxiliary graph on the Pubmed dataset in the adversarial setting.
	\label{fig:pubmed-attacked-results}}
\end{figure}

\Cref{fig:pubmed-attacked-results} shows the performance of our method using mini-batch training with a block-diagonal approximation on the \pubmed dataset in the adversarial setting. The statistics for the \pubmed dataset are shown in \Cref{tab:datasets}. We can see that this is a much larger dataset than \cora and \citeseer. In the adversarial setting, we consider attacks that add or remove edges proportionally to the total number of edges in the ground truth graph such that we remove approximately 50\% of the edges or add 50\% or 100\% of edges. For these experiments we limited the search space for the hyper-parameters to the following: \gls{GCN} regularisation in $\{ 5 \times 10^{-3}, 5 \times 10^{-4}\}$, $\smoothfactorone = \{0.25, 0.5, 0.75\}$, $\beta = \{10^{-2}, 10^{-3}\}$, $\temprior=\{ 0.5 \}$, and $\temposterior=\{ 0.5, 0.66 \}$. Finally, we reduced the given graph to $20$ non-overlapping graphs.

In \cref{fig:pubmed-attacked-results} we compare \gls{VGCN} with \gls{GCN}, \gls{GAT}, and \gls{RGCN}. We see that in the case of attacks that add edges to the graph, our \gls{VGCN} method outperforms all others. In the case of attacks that remove edges, all methods have similar performance although \gls{VGCN} displays a small performance drop and higher variance. This can easily be explained as an artifact of the block-diagonal approximation that forces \gls{VGCN} to consider a much smaller set of edges than the other methods. That is, the number of edges removed are both the attacked ones as well as the between-subgraph edges.

%\bibliography{main.bib}
%\bibliographystyle{apalike}
	
%\end{document}

\bibliographystyle{apalike}
\bibliography{main}

\end{document}